\documentclass[lettersize,journal]{IEEEtran}
\usepackage{amsmath,amsfonts}
\usepackage{algorithm}
\usepackage{array}
\usepackage{cite}
\usepackage{amsmath,amssymb,amsfonts}
\usepackage{graphicx}
\usepackage{textcomp}
\usepackage{xcolor}

\usepackage{cite}
\usepackage{amsmath,amssymb,amsfonts}
\usepackage{graphicx}
\usepackage{textcomp}
\usepackage{xcolor}
\usepackage{amssymb}
\usepackage{times}
\usepackage{soul}
\usepackage{url}
\usepackage[hidelinks]{hyperref}
\usepackage[utf8]{inputenc}
\usepackage[small]{caption}
\usepackage{graphicx}
\usepackage{amsmath}
\usepackage{amsthm}
\usepackage{booktabs}
\usepackage{algorithm}
\usepackage[switch]{lineno}
\usepackage{xcolor}
\usepackage{subfigure}
\usepackage{multirow}
\usepackage{bbm}

\usepackage{subcaption}
\usepackage{algpseudocode}
\usepackage{multirow}
\usepackage{threeparttable}
\usepackage{enumitem}
\usepackage{balance}
\usepackage{amsthm}
\usepackage{caption} 
\usepackage{bbm}
\usepackage{pifont}
\usepackage{diagbox}
\usepackage{pgfplots}
\usepackage{xcolor}
\usepackage{booktabs}
\usepackage{easyReview}
\newtheorem{definition}{Definition}

\def\BibTeX{{\rm B\kern-.05em{\sc i\kern-.025em b}\kern-.08em
    T\kern-.1667em\lower.7ex\hbox{E}\kern-.125emX}}

\newlist{enlist}{itemize}{1}
\setlist[enlist]{
  label=\textbullet\hspace{.5em},
  leftmargin=0em,
  align=left,
  labelsep=0pt,
  itemindent=1em,
  itemsep=0.2em,
  parsep=0pt,
  topsep=0.3em
}

\newlist{enenum}{enumerate}{1}
\setlist[enenum]{
label=\textup{\arabic*)\hspace{.5em}},
  leftmargin=0em,
  align=left,
  labelsep=0pt,
  itemindent=1em,
  itemsep=0.2em,
  parsep=0pt,
  topsep=0.3em
}
\begin{document}

\title{A Survey and Benchmarking of Spatial-Temporal Traffic Data Imputation Models}
\author{Shengnan Guo\textsuperscript{$\dag$}\thanks{$\dag$ These authors contributed equally.}, 
Tonglong Wei\textsuperscript{$\dag$}, 
Yiheng Huang, 
Yan Lin, 
Zekai Shen, \\
Yujuan Dong,
Junliang Lin, 
Youfang Lin, 
Huaiyu Wan\textsuperscript{*}
\thanks{* Corresponding author: Huaiyu~Wan.}
\thanks{
    \IEEEcompsocthanksitem S. Guo, T. Wei, Y. Huang, Z. Shen, Y. Dong, J. Lin are with the Key Laboratory of Big Data and 
Artificial Intelligence in Transportation, Ministry of Education, and School of Computer Science and Technology, Beijing Jiaotong University, Beijing 100044, China. \protect\\
E-mail: \{guoshn, weitonglong, huangyiheng, zkshen, dongyujuan, jlianglin\}@bjtu.edu.cn.
    \IEEEcompsocthanksitem Y. Lin is with the Department of Computer Science, Aalborg University, Aalborg 9220, Denmark. 
\protect\\
E-mail: lyan@cs.aau.dk. 
    \IEEEcompsocthanksitem Y. Lin, and H. Wan are with the Beijing Key Laboratory of Traffic Data Mining and Embodied Intelligence, School of Computer Science and Technology, Beijing Jiaotong University, Beijing 100044, China.
\protect\\
E-mail: \{yflin, hywan\}@bjtu.edu.cn.}
}



\maketitle
\begin{abstract}
Traffic data imputation is a critical preprocessing step in intelligent transportation systems, underpinning the reliability of downstream transportation services. 
Despite substantial progress in imputation models, model selection and development for practical applications remains challenging due to three key gaps: 
1) the absence of a model taxonomy for traffic data imputation to trace the technological development and highlight their distinct features.  
2) the lack of unified benchmarking pipeline for fair and reproducible model evaluation across standardized traffic datasets.  
3) insufficient in-depth analysis that jointly compare models across multiple dimensions, including effectiveness, computational efficiency and robustness. 
To this end, this paper proposes practice-oriented taxonomies for traffic data missing patterns and imputation models, systematically  cataloging real-world traffic data loss scenarios and analyzing the characteristics of existing models. We further introduce a unified benchmarking pipeline to comprehensively evaluate 11 representative models across various missing patterns and rates, assessing overall performance, performance under challenging scenarios, computational efficiency, and providing visualizations.
This work aims to provide a holistic perspective on traffic data imputation and to serve as a practical guideline for model selection and application in intelligent transportation systems. 
\end{abstract}


\section{Introduction}
\label{sec:introduction}
\begin{figure}[t]
    \centering
    \includegraphics[width=0.43
    \textwidth]{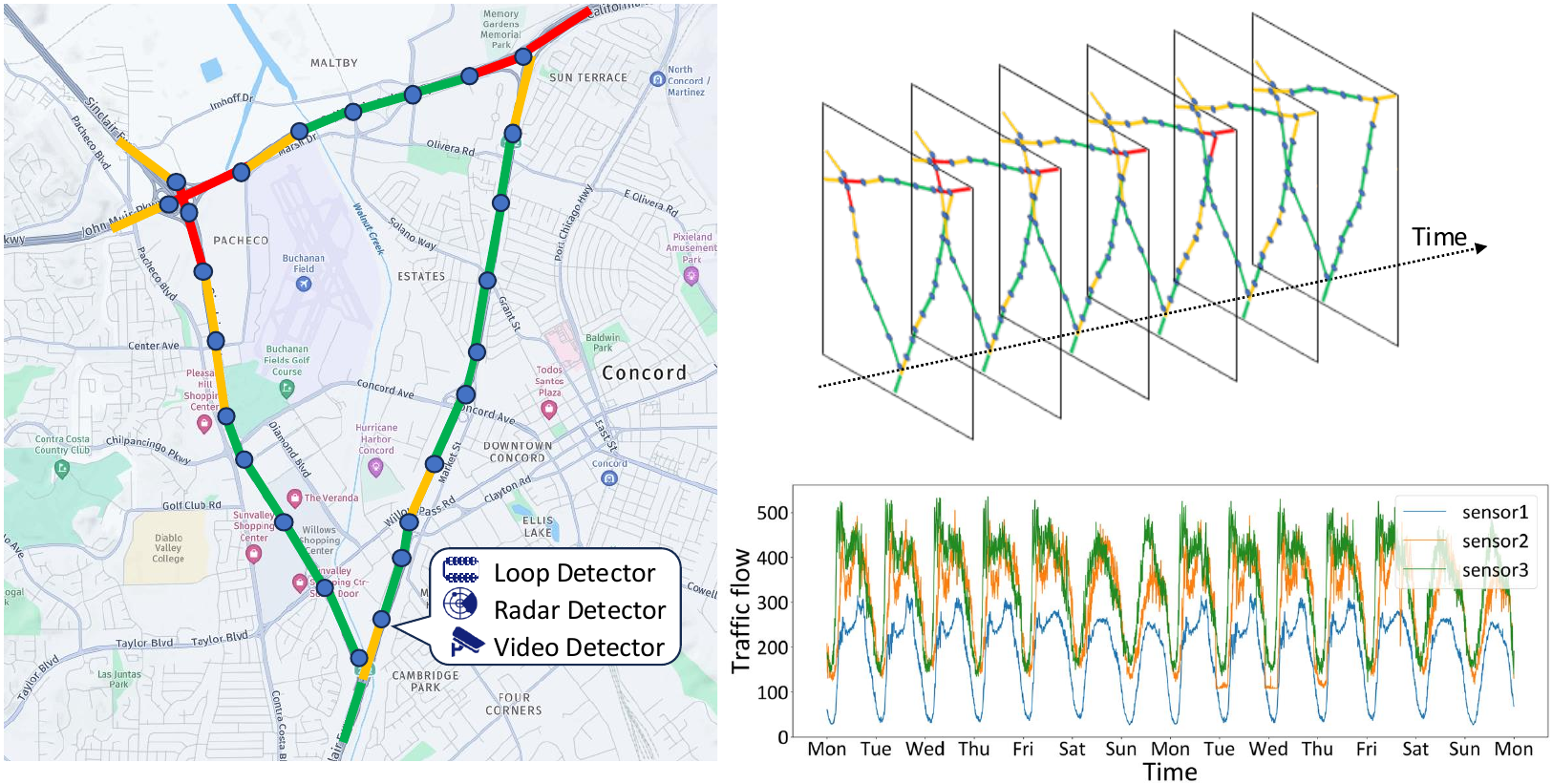}
    \caption{An illustration showing how traffic data can be modeled as spatial–temporal graph sequences.} 
    \label{fig:traffic_data}
\end{figure}

Traffic data is an example of spatial-temporal graph sequences, in which each node represents a traffic sensor that continuously collects time-series observations, and the edges represent the relationships between sensors. Fig.~\ref{fig:traffic_data} illustrates a real-world highway scenario where traffic data are collected through various sensors, such as loop detectors, radar detectors, and video detectors. These data exhibit strong spatial and temporal correlations. For example, traffic data from three connected sensors share similar changing trends, while a single traffic data sequence shows temporal autocorrelation and periodicity. 

As a fundamental resource for Intelligent Transportation Systems (ITS)~\cite{its}, it underpins key downstream services, such as traffic management~\cite{trafficmanagement}, traffic data prediction~\cite{DBLP:conf/aaai/GuoLFSW19,DBLP:conf/aaai/SongLGW20, DBLP:journals/tkde/GuoLWLC22}, and traffic signal control~\cite{DBLP:conf/nips/ChenJ0MLW24, xlight, TransformerLight}.
Despite the increasing deployment of sensors to collect real-time traffic information, missing data remains a persistent issue due to various factors like equipment malfunctions, power outages, and network failures. 
These missing data hinder ITS to accurately perceive traffic changes over time, resulting in the performance decline of downstream transportation services. 
Therefore, traffic data imputation is a crucial preprocessing step in ITS, ensuring the quality of traffic data. 

Numerous models have been specifically developed to solve this problem, focusing on two main challenges: 1) capturing the dynamics of traffic data with missing values across both spatial and temporal dimensions. 2) training models without access to ground-truth observations for the missing entries. In addition to traffic-specific imputation methods, general multivariate time series imputation models can also be adapted to this task. 
Despite significant progress in this field, developing an effective model for specific real-world scenarios in ITS remains difficult due to three key factors.

\begin{enlist}

    \item \textbf{Absence of a comprehensive survey and practical taxonomy.} 
    This gap makes it difficult to trace the technological development of traffic data imputation models and to effectively apply emerging techniques to this task. A thorough survey is needed to summarize existing models and highlight their characteristics and strengths. Moreover, a taxonomy is required to clarify the rationale behind different designs and to guide the development of new models. 
    In addition, studies on traffic forecasting~\cite{DBLP:journals/tkde/GuoLWLC22, DBLP:conf/kdd/0001L0S0LJ025, DBLP:journals/tkde/FangQLZZ24}, which are closely related to imputation and share transferable techniques, should also be incorporated and discussed. 

    \item \textbf{Lack of a unified benchmarking pipeline.} 
    Existing studies often evaluate models on different datasets and only a limited set of missing patterns. These missing patterns describe how traffic data are lost, such as randomly or continuously along spatial and temporal dimensions. However, few empirical studies compare models on standardized spatiotemporal traffic datasets with comprehensive missing patterns, making fair comparisons difficult. Additionally, although both traffic flow and traffic speed are critical for ITS, few studies have conducted experiments on both simultaneously.

    \item \textbf{Limited in-depth evaluation and analysis.} 
    First, methods evaluated within a single paper are often limited in scope; for example, newly proposed approaches from the same period or based on different technical architecture are rarely compared empirically. Second, most evaluations report only overall average errors, neglecting performance during periods with significant traffic fluctuations, which obscures model differences under challenging scenarios. Third, model efficiency, an important factor for real-world deployment, is frequently overlooked.

\end{enlist}

To address these challenges and advance research and applications in traffic data imputation, we first present a holistic overview of existing models and categorize them from a practice-oriented perspective. This provides readers with a comprehensive understanding of recent advancements and deeper insights into model characteristics. Furthermore, we design a unified benchmarking pipeline to evaluate 11 representative  models for spatio-temporal traffic data imputation, each regarded as classical within its respective technical paradigm. Our evaluation covers both traffic flow and speed datasets, encompasses diverse missing patterns across spatial and temporal dimensions, and includes detailed performance analyses. Through this, we aim to guide readers in selecting and deploying suitable models for practical scenarios.

While there are several recent surveys on data imputation models, they pay little attention to their applications in spatial-temporal traffic scenarios. For instance, Miao et al.~\cite{miao2022experimental} provides a general discussion on imputation across various data types (including numerical, categorical, and mixed data). 
Fang et al.~\cite{fang2020time} focuses on introducing deep learning-based imputation models for time series without empirical evaluations. Wang et al.~\cite{wang2024deep} and Du et al~\cite{du2024tsi} review deep learning methods for multivariate time series imputation and introduce benchmark platforms, yet they exclude imputation approaches beyond deep neural networks, such as tensor completion methods. 
Zhang et al.~\cite{zhang2024comprehensive} review public spatial-temporal traffic datasets and commonly used approaches, categorizing imputation models into interpolation-based, statistical learning-based, and prediction-based methods. However, their taxonomy is overly coarse, making it difficult to capture the commonalities and distinctive characteristics of different models. In addition, they do not provide experimental evaluations to assess which technologies most effective for different missing-data scenarios. 
Compared to them, our paper distinguishes itself in two key aspects. Firstly, we focus on the models imputation specifically applicable for spatial-temporal traffic data imputation. Secondly, beyond merely surveying these models, we conduct experiments to compare representative approaches. Rather than relying solely on overall performance metrics, we provide a detailed analysis of model behavior during critical and challenging periods, directly addressing practical concerns in the ITS domain. 

The contributions of this work are summarized as follows: 
\begin{enlist} 
    \item \textbf{Practice-oriented taxonomy.} Focusing on the task of spatial-temporal traffic data imputation, we present taxonomies for both missing patterns and imputation models. 
    Missing patterns are categorized into four types, each representing a distinct form of real-world data loss.
    Imputation models are classified based on two key aspects of practical code implementation: spatial-temporal modeling techniques and loss function design.  
    
    \item \textbf{Unified benchmarking pipeline.} We establish a  pipeline to test imputation models for spatial-temporal traffic data. The pipeline  unifies the test process including missing scenarios construction, data pre-procesing, model construction and performance evaluation. 
    Implementation of the pipeline is publicly available at \url{https://github.com/wtl52656/imputation\_benchmark}.
    
    \item \textbf{Comprehensive evaluation and in-depth analysis}. 
    We conduct extensive experiments on 11 representative models across four traffic datasets, evaluating their performance under 20  scenarios (comprising combinations of 4 missing patterns and 5 missing rates). A detailed analysis of both effectiveness and efficiency is provided to serve as a practical guideline for model selection and application. 
\end{enlist}

The rest of this paper is organized as follows:
Section~\ref{Sec:preliminaries} gives an introduction of the spatial-temporal traffic data imputation problem. 
Section~\ref{Sec:Models Summary} provides an overview of existing time-series imputation models and reviews the literature to position traffic data imputation within the broader context of related tasks. It then proposes a practice-oriented taxonomy to clarify the relationships and distinctions among these models. 
Section~\ref{Sec: Models for evaluation} introduces the imputation models evaluated in this paper. 
Section~\ref{Sec:Experiments} presents the unified benchmarking pipeline and experimental results. Section~\ref{Sec:Conclusion} concludes the paper. 


\section{Preliminaries}\label{Sec:preliminaries}
In this section, we first formulate the spatial-temporal traffic data imputation problem, then present a taxonomy on missing patterns in traffic data.  
\subsection{Problem Statement}
\begin{definition}\textbf{Traffic Network.}  We define the traffic network as a graph $G=(V, E, A)$, where $V$ represents the set of $|V|=N$ nodes (e.g., loop detectors or video cameras deployed on the traffic network). $E$ represents the set of edges that connect nodes together. $A \in {\mathbb{R}^{N \times N}}$ is the adjacency matrix representing the proximity between nodes. 
\end{definition}

\begin{definition}\textbf{Traffic Data.} The traffic data observed on node $v \in V$ at time slice $t$ is denoted as $x_{t,v} \in \mathbb{R}$. The whole traffic data observed on $G$ at time slice $t$ are defined as $\mathbf{x}_t=(x_{1,t},x_{2,t},\cdots,x_{N,t}) \in \mathbb{R}^N$. And we use $\mathbf{X}=(\mathbf{x}_1,\mathbf{x}_2,\cdots,\mathbf{x}_T) \in \mathbb{R}^{N \times T}$ to denote all observations over $T$ time slices.
\end{definition}

\begin{figure*}[h!]
    \centering
    \includegraphics[width=1\textwidth]{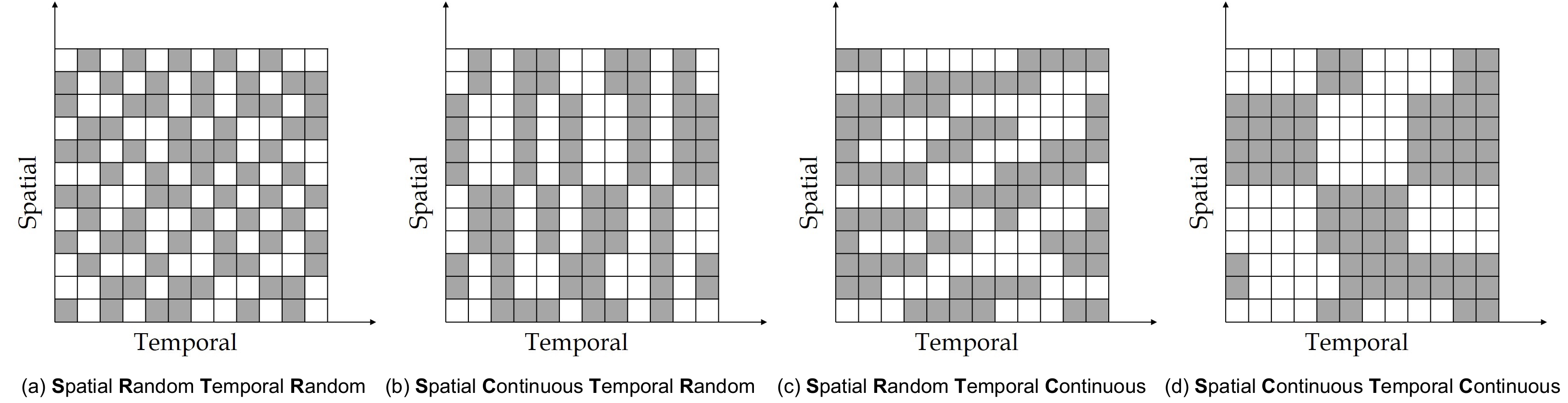}
    \caption{Illustrations of four missing patterns in traffic data, where white grids indicate missing data and gray grids indicate observed data. Specifically, SRTR may be caused by random signal or network interruption, SCTR may be caused by equipment failure over a group of devices due to some factors like network interruption in an area, SRTC may be caused by some sensors experiencing equipment failure, or network outages over a period of time, and SCTC may be caused by some reasons like power failure on a group of devices over a period of time in an area.} 
    \label{fig:missing patterns}
\end{figure*}

\begin{definition}\textbf{Masking Matrix.} To indicate the missing position in the observed traffic data, it is necessary to introduce an observation masking matrix $\mathbf{M} \in \mathbb{R}^{N \times T}$, where $m_{v,t}=0$ when $x_{v,t}$ is missing, and $m_{v,t}=1$ when $x_{v,t}$ is observed.  
\end{definition}

\begin{definition}\textbf{Time Lag Matrix.} To record the time lag between current traffic data and the last observed traffic data, a time lag matrix $\boldsymbol{\Delta} \in \mathbb{R}^{N \times T}$ is usually introduced. Each element $\delta_{v,t} \in  \boldsymbol{\Delta}$ is defined as follows,
\begin{equation}
\label{eq:timelag}
    \delta_{v,t} = \begin{cases}
     0 & \text{ if } t=1 \\
     1 & \text{ if } t>1 \text{ and } m_{v,t-1} = 1\\
     \delta_{v,t-1} + 1 & \text{ if } t>1 \text{ and } m_{v,t-1} = 0
    \end{cases}
\end{equation}
\end{definition}
\begin{definition}\textbf{Spatial-Temporal Traffic Data Imputation.} Given the incomplete observed traffic data $\mathbf{X}$ and the corresponding traffic network $G$ and masking matrix $\mathbf{M}$, the goal is to estimate complete traffic data $\hat{\mathbf{X}}$. 
\end{definition}

\subsection{Missing Patterns}
Missing patterns denote how the traffic data miss. Different missing patterns correspond to various types of real-world scenarios. This paper classifies the missing patterns in traffic data into four categories depending on the dimensions where missing happens (e.g., the spatial or temporal dimension) and 
whether the missing values appear continuously or randomly.
Specifically, Fig.~\ref{fig:missing patterns} illustrates these four missing patterns, including SRTR, SRTC, SCTR, and SCTC, where S/T indicates the Spatial/Temporal dimension, and R/C indicates missing positions are random/continuous. 

\section{Model Taxonomy}\label{Sec:Models Summary}
In this section, we start by providing a concise overview on models for time series imputation, within which traffic data imputation is a special case. 
Next, we introduce a practice-oriented taxonomy for traffic data imputation models to elucidate the relationships and distinctions among them.

\subsection{Overview on Imputation Models for Time series} 
\begin{figure}[t]
    \centering
    \includegraphics[width=0.43
    \textwidth]{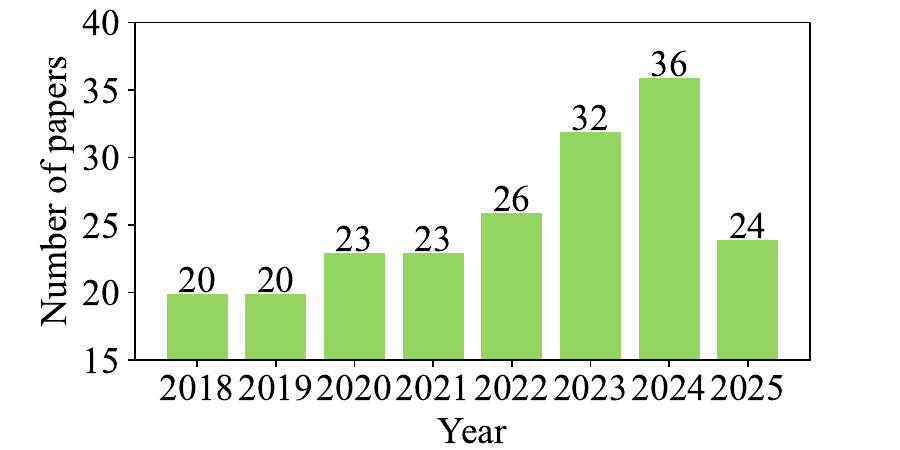}
    \caption{The annual number of papers on time series imputation published from 2018 to 2025.} 
    \label{fig:paper_sum}
\end{figure}

To provide a overall picture on research popularity for time series imputation models, we present statistics on the annual number of related published papers, as shown in Figure~\ref{fig:paper_sum}. Specifically, we search for imputation-related papers in the IEEE Xplore, ACM digital libraries, and Web of Science. Our statistics consider the papers published from 2018 to 2025 in prestigious venues, e.g., renowned conferences and journals, including ICDE, VLDB, SIGMOD, KDD, TKDE, NeurIPS, ICLR, AAAI, ICML, IJCAI, TITS. To ensure the  papers focus on time series imputation, we use keywords including ``missing data," ``imputation data," and ``interpolation" during our search.  
Finally, based on these papers, we conduct a manual filtering process to select the papers on time series imputation. 

Generally, studies on time-series imputation can be grouped into four main categories. In particular, multivariate time-series imputation is a broad area in which spatial–temporal graph data imputation is often treated as a subfield. Spatial–temporal graph imputation models represent the most suitable class of approaches for traffic data imputation. Moreover, spatial–temporal kriging can be viewed as a special case of spatial–temporal graph imputation, particularly for scenarios with continuous temporal gaps. Finally, probabilistic time-series imputation methods further estimate the uncertainty associated with the imputed values. A brief introduction to each of these four categories is given below. 

\subsubsection{Muti-variable Time Series Imputation}
Multivariate time series imputation (MTSI) has been a widely researched topic over the years, focusing on the exploitation of both temporal dependencies and feature relationships to fill in missing data. Early works utilize statistical techniques, such as the last observed value or employing $k$-nearest neighbors algorithms~\cite{troyanskaya2001missing, beretta2016nearest}. 
Tensor completion (TC) is another branch of MTSI, which harnesses the assumption of low-rank matrices to represent global correlations to impute missing data, with representative works including~\cite{khayati2020orbits, xing2023customized, xu2023hrst, fan2022dynamic,shao2018license, wang2018traffic, gong2021spatial, chen2021low}.

Within the domain of deep learning, many MTSI works have been proposed~\cite{kim2018temporal, zhang2023improving, li2022fine,luo2018multivariate,che2018hierarchical,yoon2018deep,ma2018order,zhang2019learning,wu2019hankel,jiang2020bilstm,ma2020midia,xu2021traffic,miao2021generative,luo2022neulft,zhao2022online,liu2022improving,blazquez2023selective,zhao2023transformed,li2023data}, predominantly relying on recurrent neural networks (RNNs) and Transformers to capture temporal dependence. To make the imputed data more realistic, some methods utilize adversarial training strategy~\cite{luo2019e2gan, yoon2018gain, li2019misgan,hwang2019hexagan,park2019learning,xiao2021efficient,zhang2021missing,dai2021multiple}. 
To capture both local and global temporal correlations, LGnet~\cite{tang2020joint} combines LSTMs for local dynamics with memory networks for global dynamics. NAOMI~\cite{liu2019naomi} models long-range dependencies using a multiresolution structure that recursively imputes missing values from coarse to fine resolutions via a divide-and-conquer strategy. mTAN~\cite{shukla2020multi} enhances the capture of local structures by applying the attention mechanism to latent RNN states. PrimeNet~\cite{chowdhury2023primenet} introduces an attention-based pretraining method that integrates time-sensitive contrastive learning for data imputation.

\subsubsection{Spatial-Temporal Graph Imputation}
In contrast to MTSI, spatial-temporal graph imputation (STGI)~\cite{qin2021network, wang2023traffic,li2018missing,deng2021graph,liu2023unified} incorporates an extra graph structure to explicitly represent the spatial correlations inherent in time series data. Typically, STGI utilizes recurrent neural networks (RNNs) and attention mechanisms to capture temporal dependencies, combined with graph neural networks (GNNs) to capture spatial dependencies.
For example, GRIN~\cite{cini2021filling} utilizes a specialized message-passing GNN to effectively capture spatial dependencies and impute time series in both forward and backward directions. SPIN~\cite{marisca2022learning} incorporates sparse spatial-temporal attention mechanisms, which only propagate observed data points, thus efficiently diminishing noise associated with missing values. MDGCN~\cite{liang2022memory} strengthens the modeling of spatial correlations by learning the graph structure and integrates an external memory network to retain global spatiotemporal knowledge. GCASTN~\cite{peng2023generative} introduces a generative-contrastive self-supervised learning framework, enhancing model robustness by comparing two different augmented perspectives. ImputeFormer~\cite{nie2024imputeformer} presents a Transformer-based architecture that leverages low-rankness principles to bridge the gap between deep learning and tensor completion methods for data imputation.

\subsubsection{Spatial-Temporal Kriging}
Spatial-temporal Kriging (STKriging) is designed to impute data at locations where sensors are not deployed~\cite{wang2018missing,liang2022spatial,lei2022bayesian,lao2022variational,varga2023data}, meaning the complete historical time series for these locations is unavailable.
To address this, the common approach is leveraging surrounding sensors for imputation.
KCN~\cite{appleby2020kriging} is the first deep learning-based approach for STKriging, utilizing graph neural networks (GNNs) and $k$-nearest neighbor models to impute unobserved locations by modeling spatial correlations. Building on this foundation, IGNNK~\cite{wu2021inductive} introduces a message-passing mechanism within GNNs. It is the first work to impute missing data in an inductive manner. Subsequently, many methods follow this line. 
For example, SAGCN\cite{wu2021spatial} integrates temporal dependencies using temporal convolutional networks (TCN) and spatial dependencies through a multi-aggregator mechanism. INCREASE\cite{zheng2023increase} incorporates heterogeneous spatial relations from three perspectives and leverages gated recurrent units (GRU) to model temporal correlations. DualSTN\cite{hu2023decoupling} introduces a skip graph GRU to capture long-term spatial-temporal correlations. IAGCN\cite{wei2024inductive} learns an adaptive graph structure to reconstruct spatial relationships, and STGNP~\cite{hu2023graph} incorporates uncertainty estimates for imputation.
ESC-GAN~\cite{zhang2022esc} partitions the interpolated area into grids, modeling local correlations with convolutional neural networks (CNNs) and employing a generative adversarial network (GAN) to train the model effectively.

\subsubsection{Time Series Probabilistic Imputation}
Uncertainty quantification and probabilistic modeling for traffic data ~\cite{DBLP:journals/tkde/QianZZCZZ24, DBLP:conf/icde/Qian0ZZY23} is vital in practice. 
Most imputation methods mentioned above are focused on deterministic imputation, where only a plausible value is filled in for the missing data. In contrast, Time Series Probabilistic Imputation (TSPI) seeks to provide a range for the missing values, thereby quantifying the confidence level in the imputation~\cite{yoon2018gain}. A typical approach in this domain involves using generative models to capture the data distribution. For instance, GP-VAE~\cite{fortuin2020gp} utilizes a Gaussian process as a prior to represent temporal dependencies.

Recently, diffusion models have emerged as a promising technology for TSPI. For example, CSDI~\cite{tashiro2021csdi} harnesses the stochastic characteristics of diffusion models to estimate uncertainty. It processes observed data and imputes missing values through a multi-step denoising procedure. PriSTI~\cite{Liu2023PriSTIAC} derives prior information from the available data and employs spatiotemporal attention for imputation. FastSTI~\cite{cheng2024faststi} incorporates a high-order pseudo-numerical solver to enhance inference speed. MTSCI~\cite{zhou2024mtsci} integrates a contrastive complementary mask strategy and a mixup technique into the diffusion model, leveraging conditional information from neighboring windows to ensure consistent imputation.

\begin{table*}[h]
	\centering
	\caption{Summary on the Imputation Models.}
	\centering
	\label{tab:model_summary}
	\resizebox{2\columnwidth}{!}{ 
		\begin{tabular}{cc|c|c|cc|c}
			\toprule[1pt]
			\multirow{2}{*}{Reference} & \multirow{2}{*}{Model} & \multirow{2}{*}{Tasks} & {ST Model} & \multicolumn{2}{c|}{{Loss Designs}} & \multirow{2}{*}{Dataset}\\
            \cline{5-6}
            & & & {Techniques} & {Model Types} & {Training Strategies} & \\
			\cline{1-7}
    $\bigstar$ Cao et al. - NeurIPS'18~\cite{wei2018brits}& BRITS & MTS imputation & RNN & Predictive (MAE) & Reconstruction-based SSL & Air Quality/ Health-care/ Human Activity \\
    Shao et al. - TITS'18~\cite{shao2018license}  & CTD & STG imputation & TC & Predictive  & Reconstruction-based SSL & Traffic \\
    Kim et al. - IJCAI'18~\cite{kim2018temporal} & TBM & MTS imputation & RNN & Predictive & Reconstruction-based SSL & Health-care  \\
    Luo et al. - NeurIPS'18~\cite{luo2018multivariate} & GRUI-GAN & MTS imputation & RNN & Generative (GAN) &Masked SSL & Air Quality/ Health-care \\
    Yoon et al. - ICML'18~\cite{yoon2018gain}  & GAIN & Probabilistic imputation & RNN & Generative (GAN) & Masked SSL& UCI \\
    Che et al. - ICML'18~\cite{che2018hierarchical} & MR-HDMM & MTS imputation & RNN & Generative (VAE) & Reconstruction-based SSL& Health-care/Air Quality \\
    Yoon et al. - ICLR'18~\cite{yoon2018deep} & M-RNN & MTS imputation & RNN & Predictive (RMSE) & Masked SSL & Health-care \\
    Wang et al. - TITS'19~\cite{wang2018missing} & OCC & ST Kriging & RNN & Predictive (RMSE) & Reconstruction-based SSL& Traffic \\
    Li et al. - TITS'19~\cite{li2018missing}  & MVLM & STG imputation & RNN & Predictive (MAE) &Reconstruction-based SSL & Traffic \\
    Wang et al. - TITS'19~\cite{wang2018traffic} & TAS-LR & STG imputation & TC & Predictive & Masked SSL& Traffic \\
    $\bigstar$ Luo et al. - IJCAI'19~\cite{luo2019e2gan} & E2GAN & Probabilistic imputation & RNN & Generative (GAN) & Masked SSL& Air Quality/Health-care \\
    Liu et al. - NeurIPS'19~\cite{liu2019naomi}  & NAOMI & Probabilistic imputation & RNN & Generative (GAN) &Masked SSL & Traffic \\
    Li et al. - ICLR'19~\cite{li2019misgan} & Misgan & Probabilistic imputation & RNN &  Generative (GAN) & Masked SSL& Images \\
    Ma et al. - TKDE'19~\cite{ma2018order} & OSICM & MTS imputation & KNN & Predictive (RMSE) &Reconstruction-based SSL & Air Quality/Health-care/Wiki4HE \\
    Zhang et al. - ICDE'19~\cite{zhang2019learning} & IIM & MTS imputation & KNN & Predictive (RMSE) &Reconstruction-based SSL& UCI/KEEL/Siemens \\
    Wu et al. - ICDE'19~\cite{wu2019hankel} & HKMF-T & MTS imputation & TC & Predictive (RMSE) & Reconstruction-based SSL& Traffic \\
    Park et al. - KDD'19~\cite{park2019learning} & Imp-GAIN & MTS imputation & RNN+Attention & Generative (GAN) &Reconstruction-based SSL & Health-care \\
    Jiang et al. - ICML'20~\cite{jiang2020bilstm} & BiLSTM-A & MTS imputation & RNN & Predictive & Masked SSL& Air Quality \\
    Khayati et al. - VLDB'20~\cite{khayati2020orbits} & ORBITS & MTS imputation & TC & Predictive (RMSE) &Reconstruction-based SSL & Soccer/MotionSense/BAFU/Gas \\
    $\bigstar$ Shukla et al. - ICLR'21~\cite{shukla2020multi} & mTAN & MTS imputation & RNN + Attention & Generative (VAE) &Reconstruction-based SSL & Clinical /Human Activity \\
    Gong et al. - IJCAI'21~\cite{gong2021spatial} & SMV-NMF & STG imputation & TC & Predictive (RMSE) &Reconstruction-based SSL & Urban Statistical  \\
    Xiao et al. -TITS'21~\cite{xiao2021efficient} & MTCIU GAN & MTS imputation & RNN & Generative (GAN) & Reconstruction-based SSL& Traffic \\
    $\bigstar$ Chen et al. - TITS'21~\cite{chen2021low} & LATC & MTS imputation & TC & Predictive (RMSE) & Masked SSL& Traffic \\
    Zhang et al. - TITS'21~\cite{zhang2021missing} & SA-GAIN & MTS imputation & Attention & Generative (GAN) &Masked SSL & Traffic \\
    Deng et al. - TITS'21~\cite{deng2021graph} & GTC & STG imputation & TC & Predictive (RMSE) & Reconstruction-based SSL & Traffic  \\
    Xu et al. - TITS'21~\cite{xu2021traffic} & GA-GAN & Probabilistic imputation & GNN & Generative (GAN) & Reconstruction-based SSL & Traffic  \\
    $\bigstar$ Wu et al. - AAAI'21~\cite{wu2021inductive} & IGNNK & STG imputation & GNN & Predictive (RMSE) & Reconstruction-based SSL & Traffic / Solar power \\
    Miao et al. - AAAI'21~\cite{miao2021generative} & SSGAN & MTS imputation & RNN & Generative (GAN) & Semi-Supervised Learning & Human activity/Clinic/Meteorologic \\
    Tashiro et al. - NIPS'21~\cite{tashiro2021csdi} & CSDI & Probabilistic imputation & Attention & Generative (Diffusion) & Masked SSL & Air Quality / Health  \\
    Dai et al. - ICML'21~\cite{dai2021multiple}  & MI-GAN & Probabilistic imputation & NN & Generative (GAN) & Masked SSL & ADNI/Synthetic \\
    Qin et al. - KDD'21~\cite{qin2021network} & ST-SCL & STG imputation & GNN  & Generative (VAE) & Reconstruction-based SSL & Traffic  \\
    Liang et al. - TITS'22~\cite{liang2022spatial} & STAR & ST Kriging & GNN + Attention & Predictive (RMSE) & Reconstruction-based SSL & Traffic \\
    Lei et al. - TITS'22~\cite{lei2022bayesian} & BKMF & ST Kriging & TC & Predictive (RMSE) & Reconstruction-based SSL & Health-care \\
    Wang et al. - TITS'22~\cite{wang2023traffic} & GSTAE & STG Imputation & GNN + RNN & Predictive (RMSE) & Masked SSL & Traffic  \\
    Li et al. - TKDE'22~\cite{li2022fine} & MT-CSR & MTS Imputation & CNN & Predictive (RMSE) & Reconstruction-based SSL & Traffic  \\
    Luo et al. - TKDE'22~\cite{luo2022neulft} & NeuLFT & MTS Imputation & TC & Predictive (RMSE) & Reconstruction-based SSL & Dynamic interaction networks    \\
    Fan et al. - AAAI'22~\cite{fan2022dynamic} & D-NLMC & MTS Imputation & TC  & Predictive (RMSE) & Reconstruction-based SSL &  Synthetic  Air Quality / Temperature / Chlorine Level Dataset  \\
    Marisca et al. - NIPS'22~\cite{marisca2022learning} & SPIN & MTS Imputation & GNN + Attention & Predictive (MAE) & Masked SSL &  Traffic  \\
    Ipsen et al. - ICLR'22~\cite{ipsen2022deal} & supMIWAE & Probabilistic Imputation & NN & Predictive & Supervised Learning & IMAGE / CLASSIFICATION  \\
    Liu et al. - ICLR'22~\cite{liu2022improving}  & MBMF & Probabilitisc Imputation & Causal Discovery  & Predictive (RMSE) & Reconstruction-based SSL & Synthetic / Medical  \\
    Xing et al. - TITS'23~\cite{xing2023customized} & DFCP & MTS imputation & TC & Reconstruction-based SSL & Reconstruction-based SSL & CL / LPR  \\
    Xu et al. - TITS'23~\cite{xu2023hrst} & HRST-LR & MTS imputation & TC & Predictive (RMSE) & Reconstruction-based SSL & Traffic  \\
    Blázquez-García et al. - TKDE'23~\cite{blazquez2023selective}  & MGP & MTS imputation & RNN & Predictive (RMSE) & Reconstruction-based SSL & Synthetic / CLASSIFICATION  \\
    Chowdhury et al. - AAAI'23~\cite{chowdhury2023primenet}  & PrimeNet & MTS imputation & Attention & Predictive & Masked SSL & PhysioNet / MIMIC-III  \\
    Zhang et al. - ICML'23~\cite{zhang2023improving} & UTDE-mTAND & MTS imputation & Attention & Predictive & Reconstruction-based SSL & MIMIC-III \\
   Li et al. - ICDE'23~\cite{li2023data} & BiSIM & MTS imputation & RNN + Attention & Predictive (RMSE) & Reconstruction-based SSL & Indoor Positioning \\
   $\bigstar$ Li et al. - ICDE'23~\cite{Liu2023PriSTIAC} & PriSTI & MTS imputation & GNN + Attention & Generative (Diffusion)  & Masked SSL & Air Quality / Traffic \\
    Liu et al. - VLDB'23~\cite{liu2023unified} & HMTRL & STG imputation & GNN + RNN + Attention & Predictive & Masked SSL & Traffic \\
    Li et al. - SIGMOD'23~\cite{li2023ssin} & SSIN & ST Kriging & Attention & Predictive & Masked SSL & Raingauge\\
    Chen et al. - TITS'24~\cite{chen2024low} &LRTC-SCAD&MTS imputation & TC & Predictive & Reconstruction SSL & Traffic  \\
    Zeng et al. - TITS'24~\cite{zeng2024low} &TNN-HTV&MTS imputation & TC & Predictive & Reconstruction SSL & Traffic  \\
    Yang et al. - TITS'24~\cite{yang2024latent} &LFA-TRCE&MTS imputation & TC & predictive & Reconstruction SSL & Traffic  \\
    Shu et al. - TITS'24~\cite{shu2024low}&LRTC-3DST&STG imputation & TC & predictive & Reconstruction SSL & Traffic  \\
    Wei et al. - TITS'24~\cite{wei2024self} &SAGCIN&STG imputation & Attention + GCN & Predictive(MAE) &  Masked SSL & Traffic  \\
    Li et al. - TITS'24~\cite{li2024convolutional} &CLRTR&MTS imputation & TC & predictive & Reconstruction SSL & Traffic \\
    Cheng S et al. - TITS'24~\cite{cheng2024faststi} &FastSTI&STG imputation & Attention + GCN + Diffusion Model & Generative (Diffusion) & Masked SSL & Traffic \\
    Liu et al. - TKDE'24~\cite{liu2024scope} &SGMCAI-DiT&MTS imputation & Attention + Diffusion Model & Genertaive (Diffusion) & Masked SSL & Industry  \\
    Chen et al. - TKDE'24~\cite{chen2024laplacian} &LCR-2D&MTS imputation & TC & Predictive & Reconstruction SSL & Traffic  \\
    Park Byoungwoo et al. - NeurIPS'24~\cite{park2024efficient} &PBDF&MTS imputation & Attention & Predictive(MSE) & Masked SSL & Weather / Clinical  \\
    Deng et al. - NeurIPS'24~\cite{deng2024learning} & OPCR & STG imputation & GCN + Attention &  Predictive(MAE) & Masked SSL & Traffic\\
    $\bigstar$ Nie et al. - KDD'24~\cite{nie2024imputeformer} &ImputeFormer&STG imputation & Attention  & Predictive(MAE) & Masked SSL & Traffic  \\
    Zhang et al. - KDD'24~\cite{zhang2024long} &LTVTI&MTS imputation & Attention+Diffusion  & Generative (Diffusion) & Masked SSL & AIS  \\
    Wang et al. - IJCAI'25~\cite{wang2025stamimputer} & STAMImputer & STG imputation & GCN + Attention + MoE & Predictive(MAE) & Masked SSL & Traffic \\
    Zheng et al. - TITS'25~\cite{zheng2025physics} & PRMAN & STG imputation & Attention & Predictive(MSE) & Masked SSL &  Traffic \\
    Yang et al. - AAAI'25~\cite{yang2025graph} & GSLI & STG imputation & GCN + Attention & Predictive(MSE)  & Masked SSL &  Traffic \\
    $\bigstar$ Huang et al. - AAAI'25~\cite{huang2025std} & STD-PLM & STG imputation & GCN + Attention & Predictive(MAE) & Masked SSL & Traffic \\
    Xu et al. - AAAI'25~\cite{xu2025kits} & KITS & ST Kriging & GCN & Predictive(MAE) & Masked SSL & Traffic / Air Quality \\
    Liu et al. - TKDE'25~\cite{liu2025disentangling} & TIDER & MTS imputation & TC & Predictive (MSE) & Reconstruction SSL & Traffic /  Solar power \\
    
         \midrule
			\bottomrule[1pt]
		\end{tabular}	
	}
\end{table*}

\subsection{Model Taxonomy for Traffic Data Imputation} 
\begin{figure}[ht]
    \centering
    \includegraphics[width=0.43
    \textwidth]{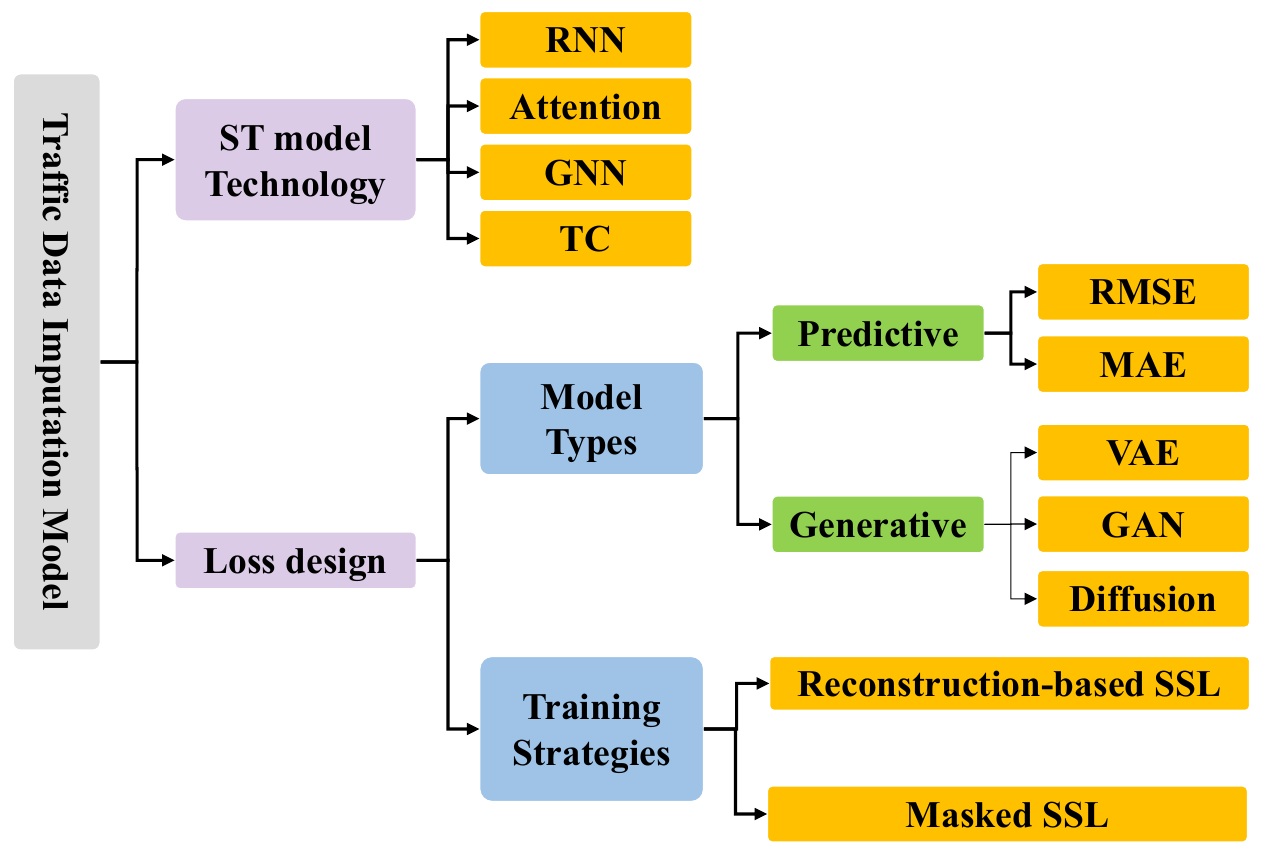}
    \caption{The practice-oriented taxonomy on imputation models applicable to traffic data.} 
    \label{fig:taxonomy}
\end{figure}

To deepen our understanding of the relationships and differences among imputation models applicable to traffic data, we propose a practice-oriented taxonomy based on two key dimensions: 
\textbf{spatial-temporal modeling techniques}, which capture inherent correlations in the data, and \textbf{loss design}, which guides model training and is tailored to the specific model type and training strategy, as shown in Figure~\ref{fig:taxonomy}. 
Both dimensions are closely aligned with practical implementation considerations. Importantly, a traffic data imputation model may combine multiple modeling techniques and adapt its loss function to suit its specific design and training paradigm. Building on this taxonomy, together with the four categories of time-series imputation tasks discussed above, we provide a comprehensive summary of representative works in Table~\ref{tab:model_summary}. 

\subsubsection{Spatial-Temporal Modeling Techniques} 
The key point in traffic data imputation is to leverage the spatial-temporal correlation of data to infer the missing values. So, in this subsection, we briefly introduce the widely employed techniques for spatial and temporal correlation learning in imputation task, including recurrent neural networks (RNN), attention mechanism, graph neural networks (GNN)~\cite{9046288}, and tensor completion (TC)~\cite{10.1145/3278607}. 
\begin{enlist}
    \item \textbf{RNN} is a sequential model commonly used to process time series data. It operates by applying the same transformation at each time step, taking the input of the current time step and the hidden state from the previous step to produce the hidden state for the current step. This sequential approach captures the temporal evolution of traffic data. Advanced variants, such as Long Short-Term Memory (LSTM)~\cite{hochreiter1997long} and Gated Recurrent Unit (GRU)~\cite{cho2020learning}, are designed to capture long-term temporal dependencies.  

 \item \textbf{Attention} offers a global receptive field, enabling the modeling of correlations between elements in a sequence regardless of their distance. This flexibility allows it to effectively capture the complex dynamics of traffic data and addresses the long-term dependency issue that RNNs struggle with. However, its quadratic space complexity demands significant computational resources.

\item \textbf{GNN} is widely used to capture the spatial correlations in graph-structured traffic data. The core design of GNNs is the pairwise message-passing mechanism, which enables nodes in the graph to iteratively update their representations by exchanging information with neighboring nodes. Representative GNN includes Graph Convolutional Network (GCN)~\cite{kipf2016semi}, Graph attention network (GAT)~\cite{velivckovic2017graph}. 

\item \textbf{TC} is a technique used to recover missing entries in a tensor. 
Matrix Completion (MC) is a special case of TC when the tensor has only two dimensions.  
TC based imputation models typically assume that the underlying tensor has a low-rank structure, optimizing it through rank minimization. This approach enables the models to capture global patterns, such as daily and weekly periodicity, in traffic data.  
\end{enlist}

\subsubsection{Loss Design} 
The design of the loss function for imputation model training is guided by two factors:  \textit{model types} and \textit{training strategies}. 

Regarding model types, there are two primary categories: \textit{predictive models} and \textit{generative models}. 
Predictive imputation models focus on learning the direct relationship between conditioned observed traffic data and the imputation target. The loss functions used for these models, such as Mean Absolute Error (MAE) and Mean Squared Error (MSE), measure the distance between the predicted output and the ground truth. On the other hand, generative imputation models aim to learn the joint probability distribution of the conditioned observations and the imputation target or to generate new data samples that align with given conditions. The loss functions for generative imputation models vary with the specific type of generative model used. Here, we introduce the most common generative models for time series imputation, including Generative Adversarial Networks (GANs)~\cite{goodfellow2014generative}, Variational Autoencoders (VAEs)~\cite{kingma2013auto}, and diffusion models~\cite{sohl2015deep,ho2020denoising,song2020score}.

\begin{enlist}
    \item \textbf{VAE} is a generative model combining deep learning and Bayesian inference. It consists of an encoder and a decoder, like a traditional autoencoder. In particular, it introduces a probabilistic latent space, encoding inputs as distributions (typically Gaussian) rather than fixed points.  
    Such probabilistic nature of VAEs allows them to model uncertainty. 
The VAE loss function, derived from the variational lower bound, consists of two parts, the reconstruction loss that measures how well the decoder reconstructs the input data from the latent space, and the KL divergence loss that ensures that the learned latent distribution stays close to a prior distribution. Formally, 
\begin{equation}
    \begin{split}
         \mathcal{L} =  \mathbb{E}_{q(z|x)}[\mathrm{log} \,p(x|z)] 
            - D_{KL}(q(z|x) || p(z)),
    \end{split}
\end{equation}
where $z$ is the latent variable and $p(z)$ is its prior distribution. $q(z|x)$ is the learned latent distribution. $p(x|z)$ represents the likelihood of the observed data $x$ given a latent variable $z$. 
    
\item \textbf{GAN} is composed of two main components: a generator $G$ and a discriminator $D$. The generator aims to transform random noise $z$ into fake data. Within the domain of data imputation, incorporating observed data into the generator's input is crucial to prevent the generation of biased imputation target. The discriminator receives both the fake data produced by the generator and truth data, then aims to assess the likelihood of each being genuine. Through adversarial training, the generator enhances its capability to produce data that closely mimics the real thing, while the discriminator strengthens its proficiency in distinguishing between authentic and synthetic data. 
The two components play a min-max game and the loss function is:  
\begin{equation}
\small
    \mathcal{L} = \min_{G} \max_{D} E_{x \sim q(x)}[\mathrm{log} \ D(x) ] + E_{z \sim p(z)}[\mathrm{log} (1 - D(G(z)))] 
\end{equation}

    \item \textbf{Diffusion model} is a type of deep generative model. It 
    begins by adding random noise to data samples over several time steps, progressively transforming them into noise, and then learns to reverse this process to recover the original data. During training, the model learns to denoise noisy data step by step, making it capable of generating new, realistic, structured samples. 
    The training objective for diffusion models typically aims to minimize the difference between the model's predicted denoised sample and the actual clean sample at each diffusion step. In practice,  the training loss is often formulated as a mean squared error (MSE) that measures the distance between the predicted noise and the actual noise added at each step in the forward diffusion process: 
    \begin{equation}
       \mathcal{L} = || \epsilon_t - \epsilon_\theta (x_t, t) ||^2,
    \end{equation}
    where $\epsilon_\theta(\cdot)$ represents the denoising network, which predicts the amount of noise that needs to be removed, and $\epsilon_t \sim \mathcal{N}(0,1)$ is the actual noise added.
    
\end{enlist}

Regarding training strategy, a critical consideration in real-world scenarios is the absence of ground truth for missing data, as past missing data cannot be retrieved. Consequently, imputation models cannot be trained against the actual ground truth. To address this, self-supervised learning (SSL) is employed. SSL methods for imputation can be further categorized by the way how observed data is used to guide model training: \textit{reconstruction-based SSL} which reconstructs observed data to guide model training, and \textit{masked SSL} which trains models by first masking portions of the observed data and then recovering them. 
Next, we delve these two training strategies.

   \begin{enlist}
    \item \textbf{Reconstruction-based SSL} strategy guides the training  of imputation model by reconstructing observed data. The corresponding loss function is usually defined as follow:
    \begin{equation}
        \mathcal{L}=\frac{1}{N}\sum_{i=1}^{N}|\mathbf{X}_i\odot \mathbf{M} - \hat{\mathbf{X}}_i\odot \mathbf{M}|,
    \end{equation}
    where $i$ is the sample index. Broadly, two types of models usually adopt this training strategy, autoencoder-based imputation models and TC-based imputation models. Specifically, autoencoder-based imputation models reconstruct the observed data by passing it through a bottleneck layer, learning compressed representations that capture meaningful patterns in the data. On the other hand, TC-based imputation models reconstruct the observed data under the assumption of a low-rank structure, effectively capturing global patterns and dependencies. 
    
    \item \textbf{Masked SSL} strategy for training imputation models is inspired by masked language modeling~\cite{devlin2018bert}. 
    Given a sample $\mathbf{X}$, the observed values in $\mathbf{X}$ are divided into two parts, then one part of them is set as the imputation targets $\mathbf{X}^{ta}$ and the other one part is set as the conditional observations $\mathbf{X}^{co}$. When training the imputation model, for each sample, the conditional observations $\mathbf{X}^{ta}$ is the input, and the imputation target $\mathbf{X}^{ta}$ is the deserved output. 
    Finally, the imputation is trained by comparing the reconstruction error with regard to the imputation target $\mathbf{X}^{ta}$. In practice, MAE is commonly chosen as the loss function. Formally, 
    \begin{equation}
        \mathcal{L}=\frac{1}{N}\sum_{i=1}^{N}|\mathbf{X}_i^{ta} - \hat{\mathbf{X}}_i^{ta}|,   
    \end{equation}
    where $i$ is the sample index.
    Masked self-supervised learning is able to simulate various missing scenarios during training by changing the choice of imputation target. Thus, the choice of imputation target is a key influencing the effectiveness of imputation in real-world application.
\end{enlist}


		

\section{Models for evaluation}\label{Sec: Models for evaluation}
In this section, we briefly introduce the models evaluated in this paper. Specifically, we establish three criteria for model selection. 1) Papers that focus on traffic data imputation. 2) Papers addressing multivariate time series imputation with high citation counts, as traffic data can be treated as a special type of multivariate time series. 3) Papers on traffic data prediction, as both prediction and imputation tasks rely on observed data to infer unobserved data. Prediction models can be adapted for imputation by modifying their training and data processing methods. Based on these criteria, we select the following 10 models for evaluation.  

\subsection{BRITS} 
BRITS~\cite{wei2018brits} is a bidirectional RNN-based model for irregular multivariate time series imputation. It is one of the earliest representatives that introduce data-driven deep-learning technologies to release existing imputation models' strong specific assumptions on the underlying data-generating process. 
BRITS uses a ``complement" input estimated by historical observations and other features at the same time $\mathbf{c}_t$ when $\mathbf{x}_t$ is missing.
Then, it introduces a temporal decay factor $\gamma_t = exp\left\{-\max(0,\textbf{W}_\gamma \delta_t + \textbf{b}_\gamma)\right\}$ to model the influence of irregular time lags $\delta_t$ on the hidden state. 
Finally, BRITS performs the computation process in both the forward and backward direction so that it can make use of both historical and future information to fill in the missing values.

\textbf{Loss design:} BRITS adopts the reconstruction-based SSL strategy for model training. 
Formally, 
\begin{equation}
    \mathcal{L} =\mathcal{L_\textit{e}}(\bf{x}_\textit{t},\hat{\bf{x}}_\textit{t})+\mathcal{L_\textit{e}}(\bf{x}_\textit{t},\hat{\bf{z}}_\textit{t})+\mathcal{L_\textit{e}}(\bf{x}_\textit{t},\hat{\bf{c}}_\textit{t}),
\end{equation}
where $\hat{\textbf{z}}_t$ is feature-based estimation, and $\hat{\textbf{c}}_t = \beta_t\odot\hat{\textbf{z}}_t+(1-\beta_t)\odot\hat{\textbf{x}}_t$, $\beta_t\in[0,1]$ used as the weight of combining $\hat{\textbf{x}}_t$ and $\hat{\textbf{z}}_t$.
$\mathcal{L}_\textit{e}$ is the mean absolute error. 

\subsection{$E^2$GAN}
E$^2$GAN~\cite{luo2019e2gan} is a GAN-based end-to-end model for multivariate time series imputation. It leverages an auto-encoder in the generator, where the encoder adds random noise $\eta$ to the input with missing values, compressing it into a low-rank latent vector $z$, the decoder then reconstructs into a complete time series. Both the encoder and decoder utilize GRUI~\cite{luo2018multivariate} to handle missing values. The discriminator consists of a GRUI layer followed by a fully connected layer to output the probability of the sample being realistic.

\textbf{Loss design:}
E$^2$GAN leverages the WGAN~\cite{arjovsky2017wasserstein} training strategy to improve stability and mitigate mode collapse. It incorporates the reconstruction-based SSL to train the generator, enabling the generation of realistic and high-quality data.
The discriminator's loss is defined as:
\begin{equation}
    \mathcal{L}_D = -D(\mathbf{X}) + D(\hat{\mathbf{X}})
\end{equation}
To impute missing data, the generator's loss incorporates a mean squared error:
\begin{equation}
    \mathcal{L}_G = \lambda \| \mathbf{X} \odot \mathbf{M} - G(\mathbf{X} + \eta) \odot \mathbf{M} \|_2^2 - D(\hat{\mathbf{X}})
\end{equation}
To strengthen the generator, E$^2$GAN updates the discriminator every $k$ iteration of generator updates during training.

\subsection{mTAN}
mTAN~\cite{shukla2021multi} is a VAE-based time series imputation model. By combining attention and VAE, the sparse and irregularly sampled time series can be projected into a latent space. 
Unlike conventional attention, mTAN calculates attention scores by time series and a series of reference points.
The encoder of mTAN projects irregularly sampled time series data into a latent space of fixed dimension through specially designed attention, and the decoder recovers missing data based on the attention between the latent vector and reference points.

\textbf{Loss design:}  mTAN employs the reconstruction-based SSL strategy to train the model. The learning objective of mTANs is shown below, with some minor changes from the VAE, 
\begin{equation}
    \begin{split} 
    \mathcal{L}_{\mathrm{NVAE}}(\theta, \gamma)=
    \sum_{t=1}^{T} \frac{1}{N}
    (\mathbb{E}_{q_{\gamma}\left(\mathbf{z} \mid \mathbf{r}, \mathbf{x}_{t}\right)}\left[\log p_{\theta}\left(\hat{\mathbf{x}}_{t} \mid \mathbf{z}, t \right)\right]  \\
    -D_{\mathrm{KL}}(q_{\gamma}\left(\mathbf{z} \mid \mathbf{r}, \mathbf{x}_{t}\right) \| p(\mathbf{z}))),
\end{split}
\end{equation}
where $\mathbf{z}$ is the defined latent variable assumed to follow a multidimensional normal distribution $p(\mathbf{z})$. $q_{\gamma}$ is the distribution obtained by the encoder, $p_{\theta}\left(\hat{\mathbf{x}}_{t} \mid \mathbf{z}, t\right)$ refers to the probability of the decoder obtaining $\hat{\mathbf{x}}_t$ based on $\mathbf{z}$ and $t$.

\subsection{IGNNK}
IGNNK\cite{wu2021inductive} focuses on spatial-temporal kriging task, whose structure consists of three graph neural network layers. The most important design of IGNNK is that it can use the message passing mechanism of the graph neural network to aggregate the information of neighbor nodes inductively so as to be applied to other datasets with different graph structures without retraining.

\textbf{Loss design:} IGNNK implements the masked SSL strategy to train the model, yet it simultaneously evaluates observed and missing data.
To make the model inductive for unseen graphs, it randomly samples subgraphs in each batch to simulate different graph structures. At each subgraph, IGNNK randomly selects some nodes as the sampled nodes and the others as unsampled nodes. To make the message passing mechanism more generalized for all nodes, IGNNK defines the loss function as follows, which uses the total reconstruction error on both observed and unseen nodes:
\begin{equation}
    \mathcal{L}=\sum\left\|\hat{\mathbf{X}}-\mathbf{X}\right\|^2
\end{equation}


\subsection{LATC}
LATC \cite{chen2021low} is a low-rank autoregressive tensor completion model designed for spatial-temporal traffic data. The model initially reshapes the traffic data matrix $\mathbf{X} \in \mathbb{R}^{N \times T}$ into a third-order tensor \(\mathcal{X} \in \mathbb{R}^{N \times I \times J}\) to capture the periodic of traffic data, where \(I\) represents the number of time slots within a day, \(J\) represents the number of days, and \(I \times J = T\). 
Subsequently, LATC follows to the low-rank assumption, aiming to minimize the tensor rank to effectively capture the global patterns.
Meanwhile, LATC introduces a temporal variation term as a new regularization component to capture local temporal consistency. The temporal variation is defined as the cumulative sum of autoregressive errors within the unfolded time series matrix. To approximate the low-rank characteristic and circumvent the challenges associated with determining the rank in factorization models, LATC employs the truncated nuclear norm of the completed tensor.

\textbf{Loss design:}  
It introduces an autoregressive norm  $\left\|\hat{X}\right\|_{W,{\mathcal{H}}}$ to capture local trends within the data. This autoregressive norm is defined as follows, which is a combination of the reconstruction-based SSL loss function and autoregression:
\begin{equation}
    \left\|\hat{\mathbf{X}}\right\|_{\mathbf{W},{\mathcal{H}}}= \sum_{n=1}^{N}\sum_{t=1}^{T} [\hat{\mathbf{X}}_{n,t}- (\sum_i^d \mathbf{W}_{n,i}\hat{\mathbf{X}}_{n,t-h_i} )]^2,
\end{equation}
where \(d\) is a hyperparameter to determine the number of historical values, \(h_i\) represents the time interval between the time of the \(i\)-th historical value and the current time \(t\), \(\mathcal{H} = \{h_1, \cdots, h_d\}\) denotes the set of these time intervals, and \(\mathbf{W} \in \mathbb{R}^{N \times d}\) is the coefficient matrix that models the contribution of each historical value to the current time step.

Additionally, LATC preserves global trends in the imputed data by introducing a forward tensorization operator \(\mathcal{Q}(\cdot)\) that transforms the imputed traffic matrix \(\hat{\mathbf{X}}\) into a third-order tensor \(\hat{\mathcal{X}} = \mathcal{Q}(\hat{\mathbf{X}})\). To ensure the low-rank characteristic of the tensor and capture the global patterns, LATC minimizes the nuclear norm of the third-order tensor \(\|\hat{\mathcal{X}}\|_{r,*}\). Formally, 
\begin{equation}
\begin{aligned}
&\min_{\hat{\mathcal{X}},\hat{\mathbf{X}},W}\left\|\hat{\mathcal{X}}_{r,*}\right\|+\frac{\lambda}{2}\left\|\hat{\mathbf{X}}_{W,\mathcal{H}}\right\|, \\
    &\text{s.t.} \text{   } \hat{\mathcal{X}} =   \mathcal{Q}(\hat{\mathbf{X}}), \mathcal{P}_{\Omega}(\hat{\mathbf{X}}) =  \mathcal{P}_{\Omega}(\mathbf{X}),
\end{aligned}
\end{equation}
where the operation \(\mathcal{P}_{\Omega}(\cdot) \) retains the observable values while setting the missing positions to \(0\), and \(r\) is a truncation parameter.

\subsection{PriSTI}
PriSTI~\cite{Liu2023PriSTIAC} is a self-supervised spatiotemporal data imputation model based on diffusion. The model consists of two components: the Conditional Feature Extraction Module (CFEM) and the Noise Estimation Module (NEM). CFEM learns spatiotemporal prior first linearly impute data, then using graph neural networks and attention mechanisms. NEM serves as the noise prediction module within the diffusion framework, employing both temporal and spatial attention to capture spatiotemporal correlations and geographical relationships. CFEM provides NEM with a better global contextual prior knowledge, aiding in its further understanding of spatiotemporal dependencies. During the reverse denoise process, CFEM and NEM progressively recover real data from the noise by multi-step denoise.

\textbf{Loss design:} PriSTI adopts diffusion loss with the masked SSL strategy to evaluate the added noise on missing data.
During training, it only computes the error at the locations where missing values are artificially introduced. The loss function is similar to the conventional diffusion model.

\subsection{GCASTN}
GCASTN is a spatial-temporal imputation model that utilizes generative-contrastive learning. It constructs two views to reconstruct the complete data using the same encoder and decoder. In the encoder, GCASTN introduces the $\delta $ to represent the time interval between the current missing position and its closest observable position in the past, then adaptively adjusts the learned spatial-temporal dependencies to make the imputation more reliable.

\textbf{Loss design:} GCASTN adds contrastive learning based on the masked SSL, wherein its loss function comprises two parts. The first is reconstructing the missing values in two views:
\begin{equation}
    \mathcal{L}_G = \sum (\left \| \hat{\mathbf{X} }_1 - {\mathbf{X}} \right \|_2^2 
+ \left \| \hat{\mathbf{X} }_2 - {\mathbf{X}} \right \|_2^2  )
\end{equation}
The second loss is the contrastive loss that aligns different views:
\begin{equation}
    \mathcal{L}_C = \sum (\left \| \hat{\mathbf{X} }_1 - \hat{\mathbf{X}}_2 \right \|_2^2),
\end{equation}
where $\hat{\mathbf{X} }_1$ and $\hat{\mathbf{X} }_2$ are two positive data views. The total loss $\mathcal{L} = \mathcal{L}_G + \alpha \mathcal{L}_C$ by a weight $\alpha$.

\subsection{AGCRN}
AGCRN is a classic graph neural network for traffic prediction. It designs a novel data-driven node embedding, and then implements adaptive GCN through node embedding, thus getting rid of the constraints of prior graph structure. In addition, it reduces the complexity of GCN through matrix decomposition operations while providing an independent parameter space for each node.

\textbf{Loss design:} 
The original AGCRN uses the mean absolute error to predict future flow. Here, we train it using the masked SSL strategy, making it suitable for imputation tasks.

\subsection{ASTGNN}
ASTGNN is an attention-based spatiotemporal graph neural network designed for traffic forecasting.  It utilizes the same sequence-to-sequence structure as the Transformer. 
To model the local trends within the data, ASTGNN introduces a novel trend-aware attention mechanism that discovers local change trends from the local context.
In summary, ASTGNN can effectively capture both global and local spatiotemporal dependencies within the data, thereby enabling accurate forecasting of spatiotemporal data.

\textbf{Loss design:} To enable ASTGNN for the imputation task, we adopt the same  masked SSL strategy as AGCRN.

\subsection{ImputeFormer}
ImputeFormer~\cite{nie2024imputeformer} combines the Transformer with low-rank induction to have both prior structure and representation capabilities for spatiotemporal imputation. It balances inductive bias and model expressivity to handle missing data in various settings. The model primarily incorporates attention mechanisms in both the temporal and spatial dimensions, with a special design imposing low-rank constraints on the two attention operations. Specifically, for temporal attention, ImputeFormer applies low-rank constraints by projecting the data into a low-dimensional space. For spatial attention, it replaces the traditional attention score computation with inner products of low-dimensional node embeddings, thereby imposing low-rank constraints. 

\textbf{Loss design:} ImputeFormer utilizes masked SSL to train the model. Its loss function integrates MAE loss $\mathcal{L}_{\mathrm{recon}}$ to reconstruct the masked part. It also introduces a Fourier Sparsity Regularization $\mathcal{L}_{\mathrm{FIL}}$ to maintain low-rank structures by imposing spectral constraints on imputations. 
The loss function is defined as follows:
\begin{equation}
\begin{split}
    \mathcal{L}_{\mathrm{recon}} & =\sum\|\mathbf{M} \odot(\widehat{\mathbf{X}}-\mathbf{X})\|_1, \\
    \bar{\mathbf{X}} &=\mathbf{M}\odot\mathbf{X} + (\mathbf{1-M}) \odot \widehat{\mathbf{X}}, \\
    \mathcal{L}_{\mathrm{FIL}} & =\sum\|\mathrm{Flatten}(\mathrm{FFT}(\bar{\mathbf{X}})\|_{1}, 
\end{split}
\end{equation}
where FFT means the Fast Fourier Transform.

\subsection{STD-PLM}
STD-PLM~\cite{huang2025std} is a unified spatial-temporal prediction and imputation model built upon pre-trained language models. By converting spatial-temporal data into token sequences through a specially designed spatial-temporal tokenizer, STD-PLM leverages the powerful comprehension and reasoning capabilities of pre-trained language models for both forecasting and imputation tasks. To enhance the model’s awareness of the spatial topological structure inherent in spatial-temporal data, STD-PLM introduces an adaptive node embedding mechanism based on the adjacency matrix and a region-level token construction strategy. These designs jointly enable STD-PLM to achieve competitive performance on both prediction and imputation benchmarks.

\textbf{Loss design:} Following PriSTI, STD-PLM adopts a masked SSL strategy, wherein additional conditional missingness is artificially injected into the training data to construct supervision samples.

\section{Experiments}\label{Sec:Experiments}
\subsection{Datasets}
We evaluate the representative imputation models mentioned above using public datasets on two kinds of traffic imputation tasks, including highway traffic flow imputation and highway traffic speed imputation. The detailed statistical information of datasets is summarized in Table~\ref{tab:dataset}.
\subsubsection{Highway Traffic Flow}
In the highway traffic flow imputation task, we conduct experiments on two commonly used datasets~\cite{guo2019attention}, namely $\textbf{PEMS04}$ and $\textbf{PEMS08}$, which are collected by the Caltrans Performance Measurement System (PeMS) and the traffic flow is aggregated into 5-minute windows. \textbf{PEMS04} records 307 sensors traffic flow from January 1st, 2018 to February 28th, 2018, and \textbf{PEMS08} records 170 sensors traffic flow from July 1st, 2016 to August 31st, 2016.
\subsubsection{Highway Traffic Speed}
We choose \textbf{Seattle} dataset~\cite{cui2019traffic} to evaluate the highway traffic speed imputation task. This dataset is collected in Seattle area, which records the speed of 323 loop detectors from January 1st, 2015 to December 31st, 2015 and the time interval is 1 hour.

\subsubsection{TW}
We also collected a new dataset TW from public sources 
for our experiments. The TW dataset comprises traffic flow data from 315 sensors across Taiwan, recorded at 5-minute intervals, spanning the period from November 1st, 2020 to November 21st, 2020. 

\begin{table}[htb]
	\renewcommand{\arraystretch}{1.3}
	\caption{Dataset Description.}
	\resizebox{1\columnwidth}{!}{ 
		\label{tab:dataset}
		\centering
		\begin{tabular}{ccccc}
			\toprule
			Data type &Datasets & \# Nodes & Time Range &  Time interval\\
			\hline
			\multirow{2}{*}{Traffic Flow}&PEMS04 & 307 & 01/01/2018 - 02/28/2018 & 5 minutes\\
			& PEMS08 & 170 & 07/01/2016 - 08/31/2016 & 5 minutes\\
                & TW & 315 & 11/01/2020 - 11/21/2020 & 5 minutes\\
			\hline
			Traffic Speed & Seattle & 323 & 01/01/2015 - 12/31/2015 & 1 hour\\
			\bottomrule
		\end{tabular}
	}
\end{table}

\begin{figure*}[!h]
	\centering
	\subfigure{		\includegraphics[width=0.233\linewidth]{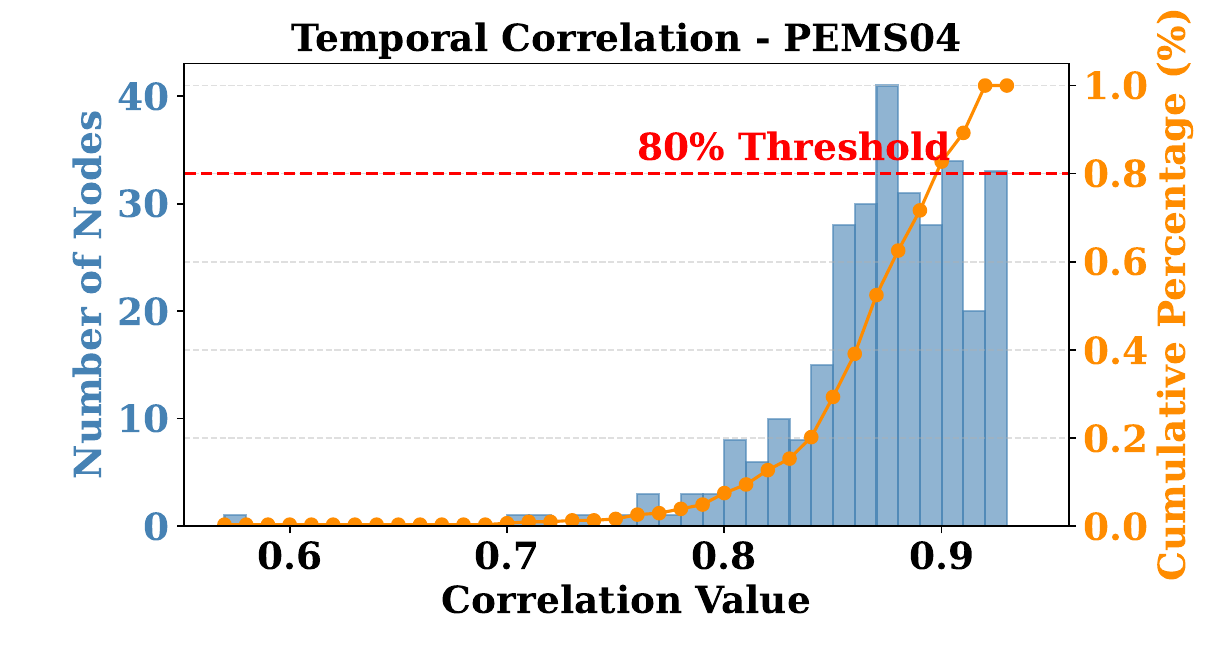}}
    \hspace{0.mm} 
	\subfigure{\includegraphics[width=0.233\linewidth]{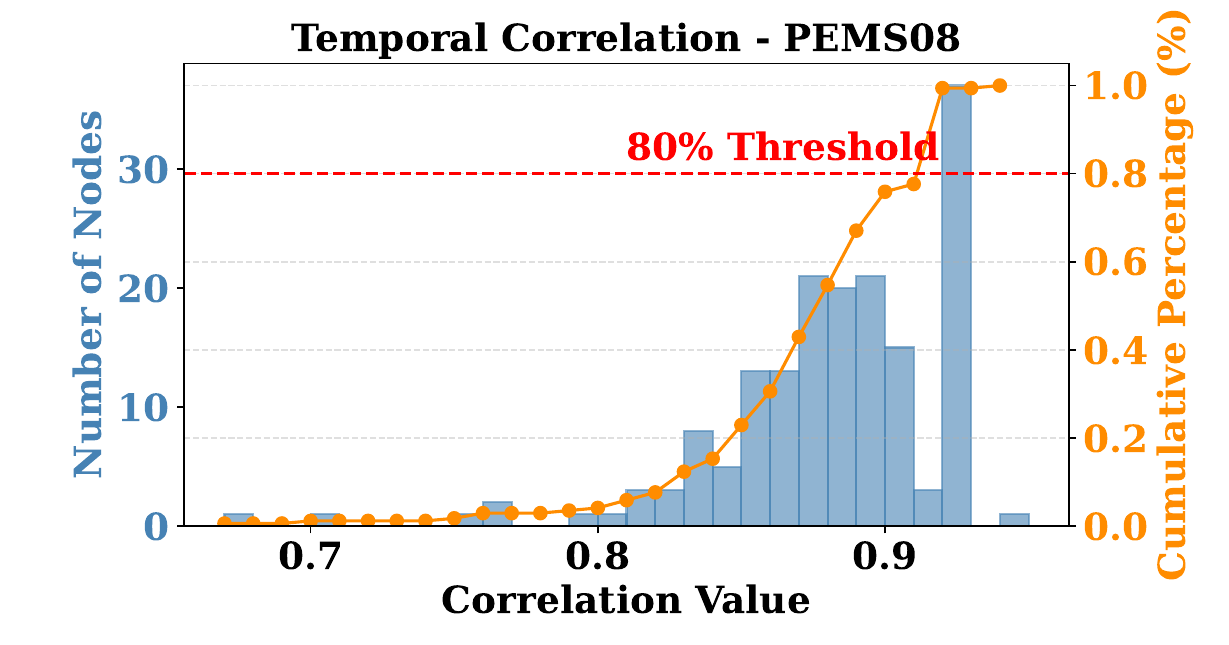}}
    \hspace{0mm} 
        \subfigure{		\includegraphics[width=0.233\linewidth]{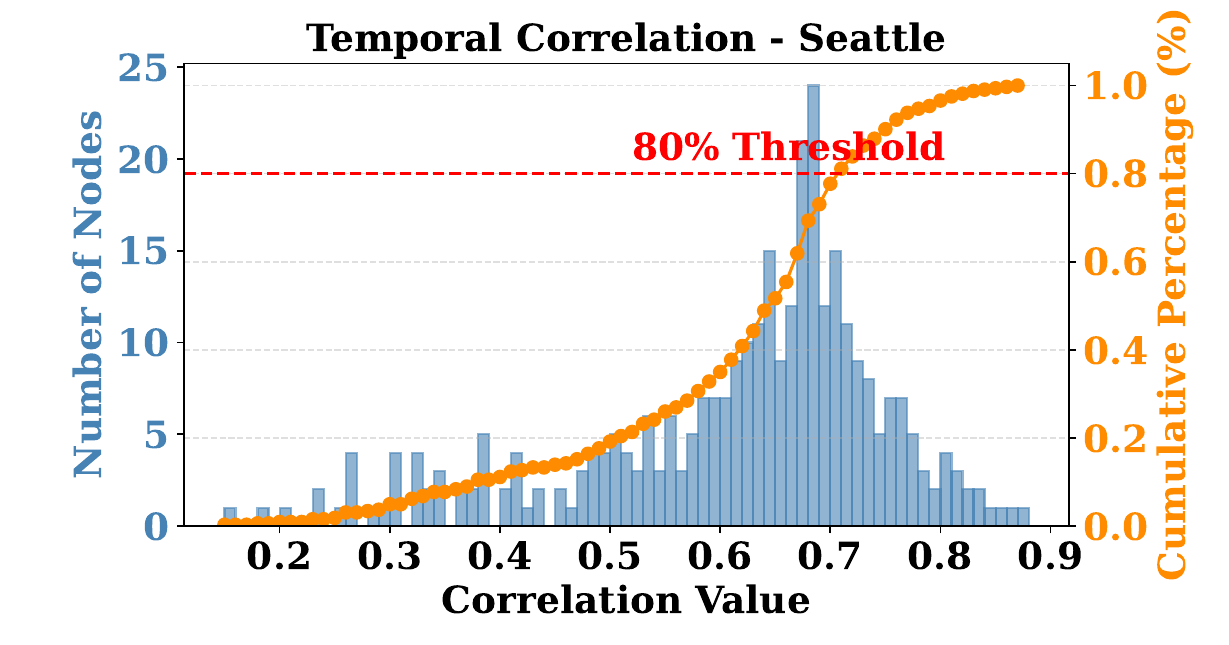}}
        \hspace{0.mm} 
	\subfigure{\includegraphics[width=0.233\linewidth]{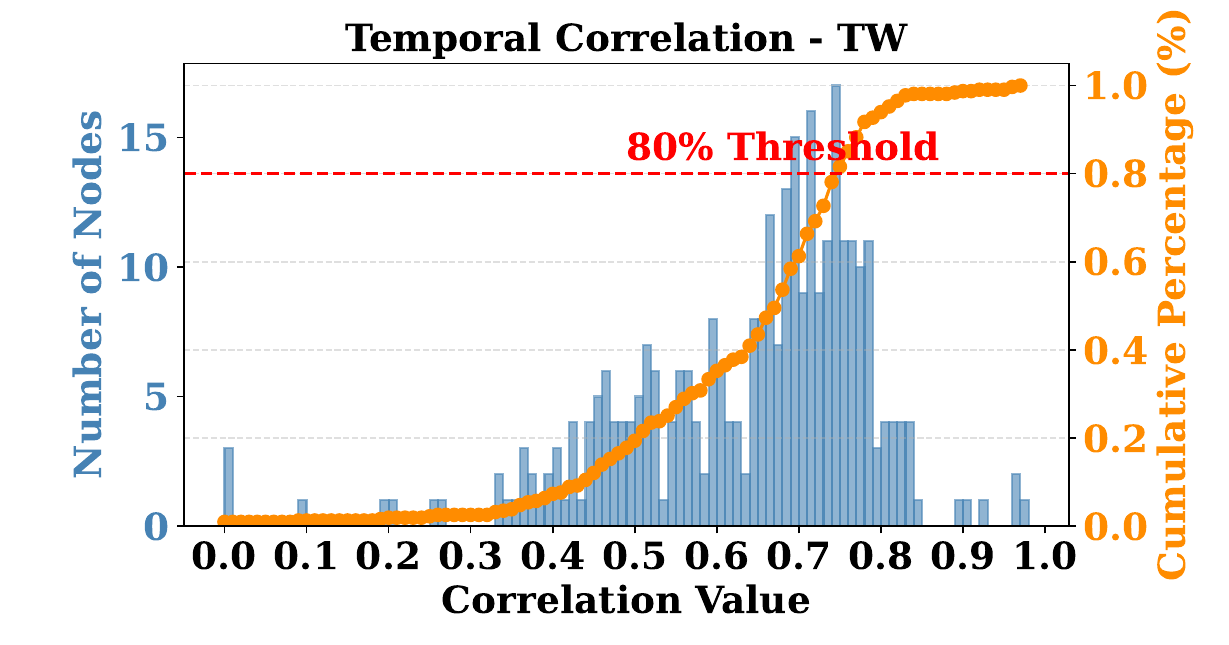}}
    \hspace{0.mm} 
    \subfigure{		\includegraphics[width=0.233\linewidth]{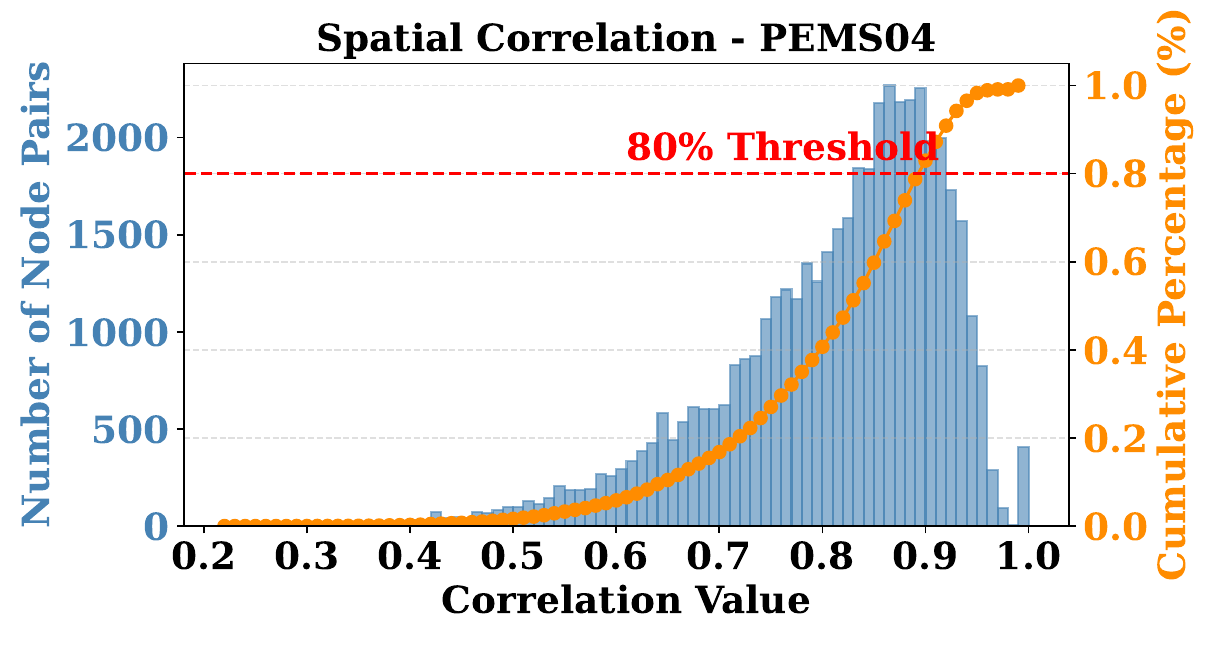}}
    \hspace{0.mm} 
	\subfigure{\includegraphics[width=0.233\linewidth]{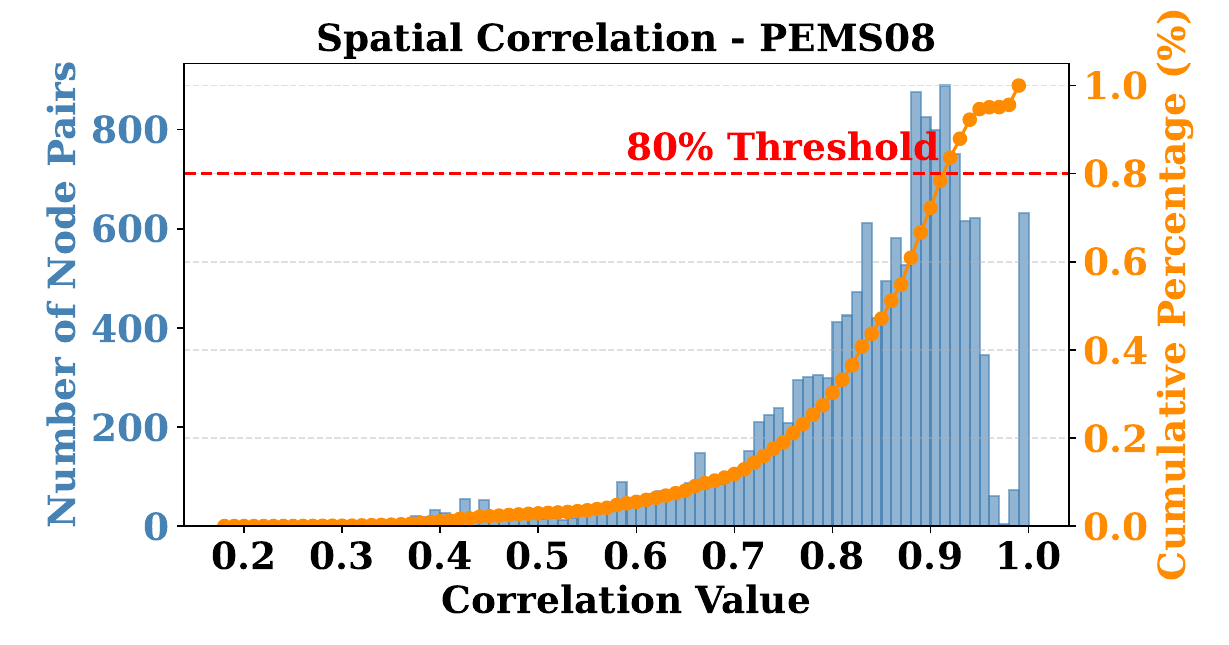}}
    \hspace{0mm} 
        \subfigure{		\includegraphics[width=0.233\linewidth]{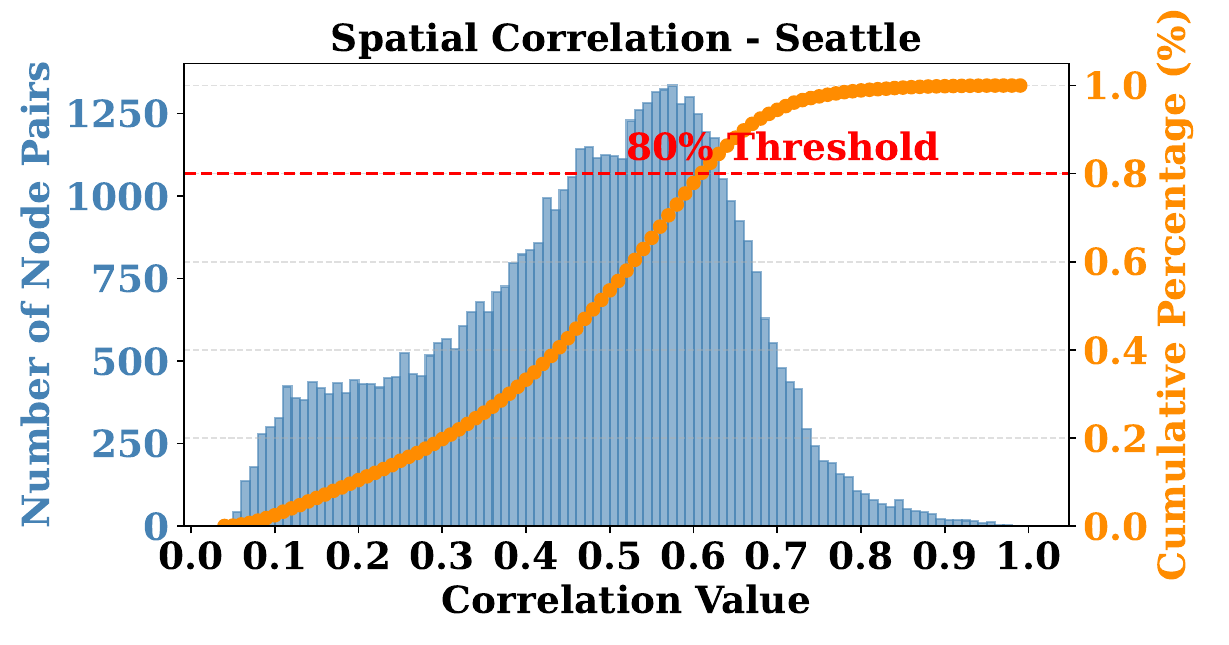}}
        \hspace{0.mm} 
	\subfigure{\includegraphics[width=0.233\linewidth]{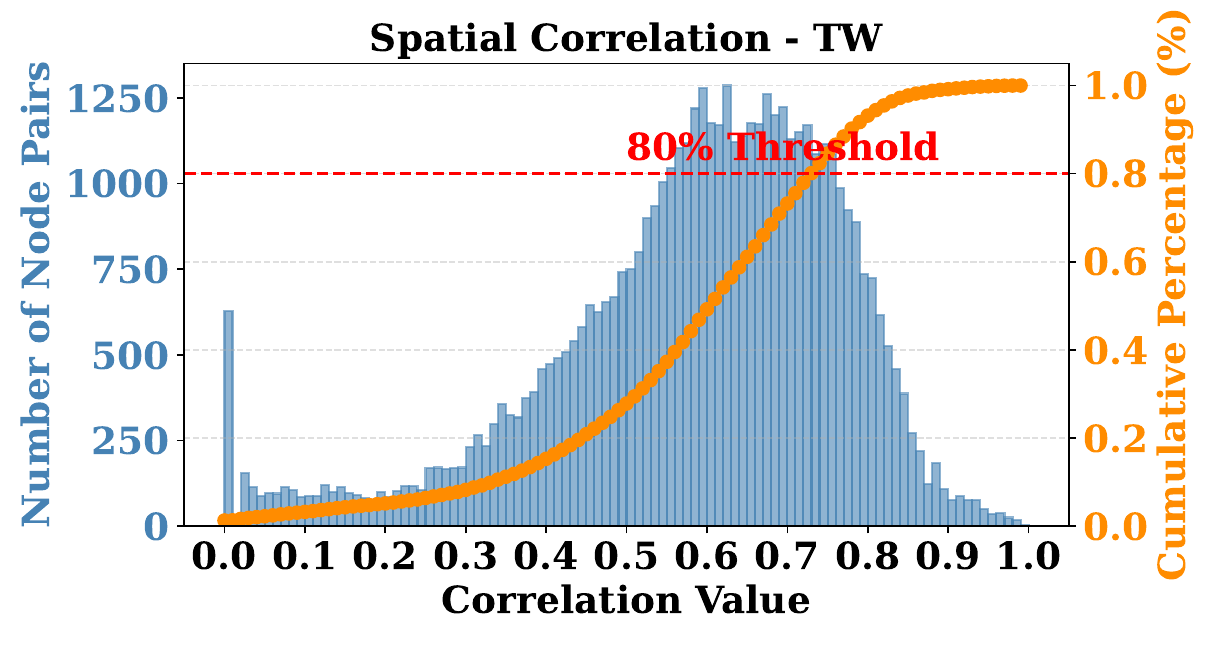}}
	\caption{Spatiotemporal Correlation of different datasets.}
	\label{fig:st_correlation}
\end{figure*}

\textbf{Dataset Spatiotemporal Correlation Analysis.} As shown in Figure~\ref{fig:st_correlation}, we conduct a dataset-level macro spatiotemporal correlation analysis. For temporal correlation, we consider the traffic flow/speed for each station as a time series and calculate the dynamic time warping (DTW) similarity between time series from one day to the following day. For spatial correlation, we use traffic flow data for one day at each node, compute the DTW across $N$ sequences, and analyze their similarity.
The experimental results are illustrated in the figure. We observe that PEMS04 and PEMS08 show stronger temporal correlations, with {95.11\% and 96.47\%} of the data reaching a temporal similarity of 80\%. In contrast, only 7.30\% of TW's data reached 80\% similarity, and the Seattle dataset only has 4.64\% data. 

{As for spatial correlation, PEMS04 and PEMS08 demonstrate stronger results, reaching 80\% similar with {62.31\% and 72.62\%} data, respectively. In Seattle, only {1.30\%} of nodes achieve 80\% similarity. In TW, {8.29\%} of nodes reached 80\% similarity.}

\begin{figure*}[t]
    \centering
    \includegraphics[width=1\textwidth]{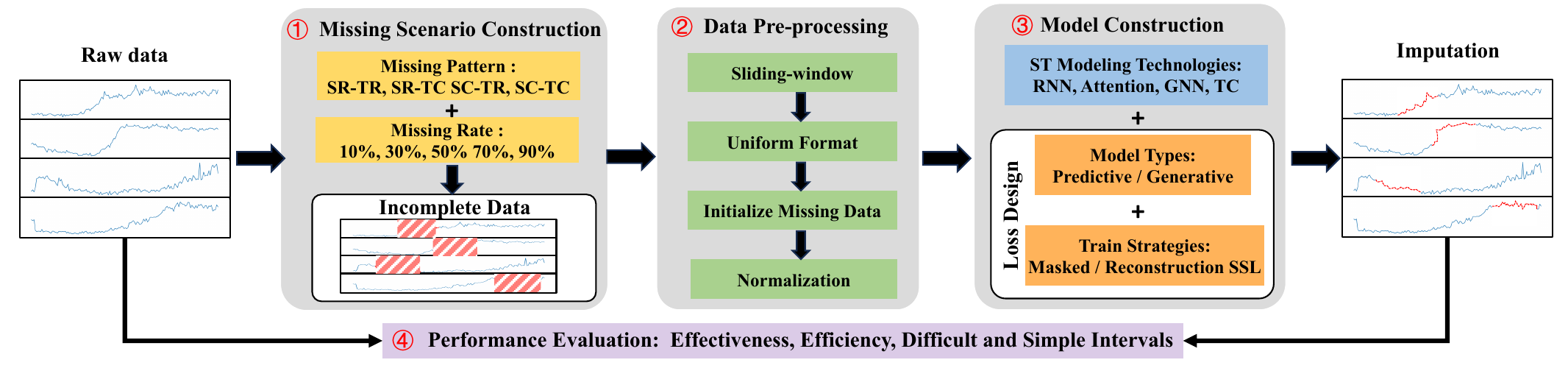}
    \caption{The illustration of the pipeline, which consists of missing scenario construction, data pre-processing, model construction, and performance evaluation.}
    \label{fig:pipeline}
\end{figure*}
\subsection{Evaluation Pipeline}
We propose a unified evaluation pipeline for traffic data imputation, as illustrated in Figure~\ref{fig:pipeline}. This pipeline standardizes the whole process of evaluating imputation models by four steps, including missing scenario construction, data pre-processing, model construction, and evaluation.
\subsubsection{Missing Scenario Construction} 
We construct 20 missing scenarios based on real-world conditions, including four distinct missing patterns, each with five missing rates $\alpha$. The construction process of each missing pattern is as follows:
\begin{enlist}
    \item \textbf{SRTR.} We randomly generate a mask matrix $\mathcal{M}$ with the same dimensions as the original data $\mathcal{X}$. Each element in $\mathcal{M}$ is assigned a value of 0 with a probability of $\alpha$, and a value of 1 with a remaining probability of $1-\alpha$.
    
    \item \textbf{SRTC.} In this missing pattern, we initially define a continuous time splice of length $l_m$ as a patch and divide the dataset across the time dimension into $\frac{T}{l_m}$ non-overlapping patches. We then introduce missing scenarios at the patch level, randomly masking $\frac{T}{l_m} * N *\alpha$ patches.
    \item \textbf{SCTR.} For the spatial continuous missing, 
    we employ a graph clustering algorithm~\cite{blondel2008fast} to group all sensors into $N_c$ clusters based on distance. And each sensor is assigned a cluster. We construct missing scenarios at the cluster level, randomly masking out $T * N_c * \alpha$ clusters.
    \item \textbf{SCTC.} We initially generate $\frac{T}{l_m} * N_c$ non-overlapping blocks. Then we introduce missing scenarios at the block level, randomly masking out $\frac{T}{l_m} * N_c * \alpha$ blocks.
     
\end{enlist}

\subsubsection{Data Pre-process}



The dataset undergoes a three-step pre-processing procedure. Initially, we establish the sliding window size of $T_w$, to generate data samples for model training. For the LATC model, all training samples are input simultaneously. Next, acknowledging that many traditional imputation models are specifically designed for multi-variable time series without spatial dimensions, we reshape the data samples to align with each model's input format. Concurrently, we assign a value of 0 to the missing data locations. Lastly, we apply the normalization method detailed in the original paper to standardize the data. 

\subsubsection{Model Construction}
We summarize 10 recently proposed sequence imputation and prediction models. For constructing the model, we follow the three steps: First, we adapt the code from the official release and only adjust the data input and output to align with traffic data. Secondly, we conduct a large-scale grid search to optimize their hyperparameters for peak performance. Finally, we train each model using a consistent training dataset, guided by a loss function, with an early stopping criterion applied to the validation dataset.

\subsubsection{Performance Evaluation} 
We conduct a detailed evaluation of 20 missing scenarios and 10 baselines across three key aspects. For effectiveness, we use three commonly adopted metrics to assess mean error: Root Mean Square Error (RMSE), Mean Absolute Error (MAE), and Mean Absolute Percentage Error (MAPE). For efficiency, we compare their training time, inference time, and memory usage. Additionally, we evaluate their imputation performance across both challenging and stable periods to simulate varying real-world traffic conditions.
Note that our evaluation only focuses on the performance at the locations where data is missing.




\subsection{Effectiveness  and Robustness Evaluations}
\subsubsection{Imputation Performance on Different Datasets}

\begin{figure*}
    \centering
    \includegraphics[width=0.95\linewidth]{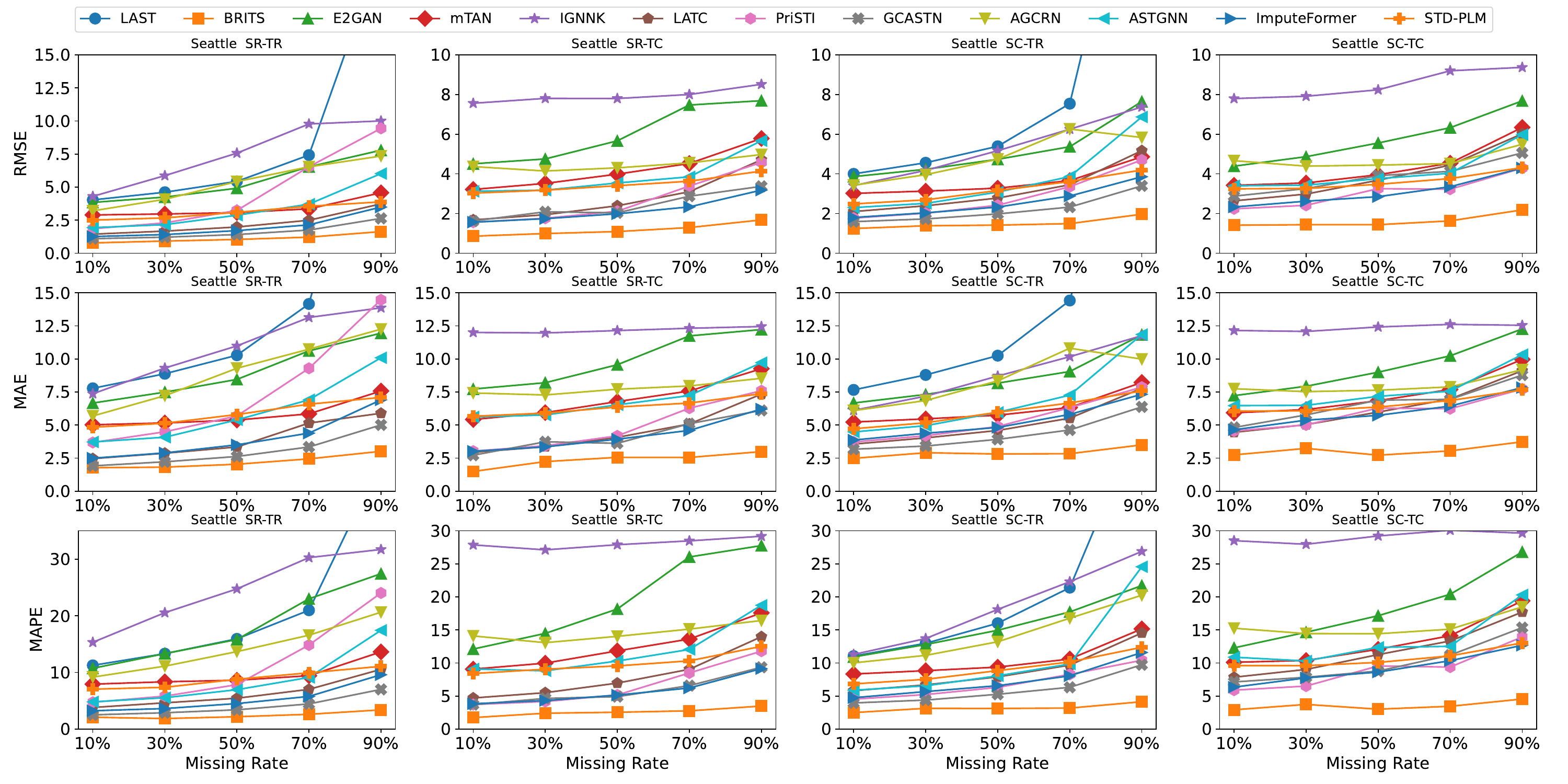}
    \caption{Performance of traffic data imputation for different models under 20 missing scenarios at Seattle.}
    \label{fig:Seattle}
\end{figure*}

\begin{figure*}
    \centering
    \includegraphics[width=0.95\linewidth]{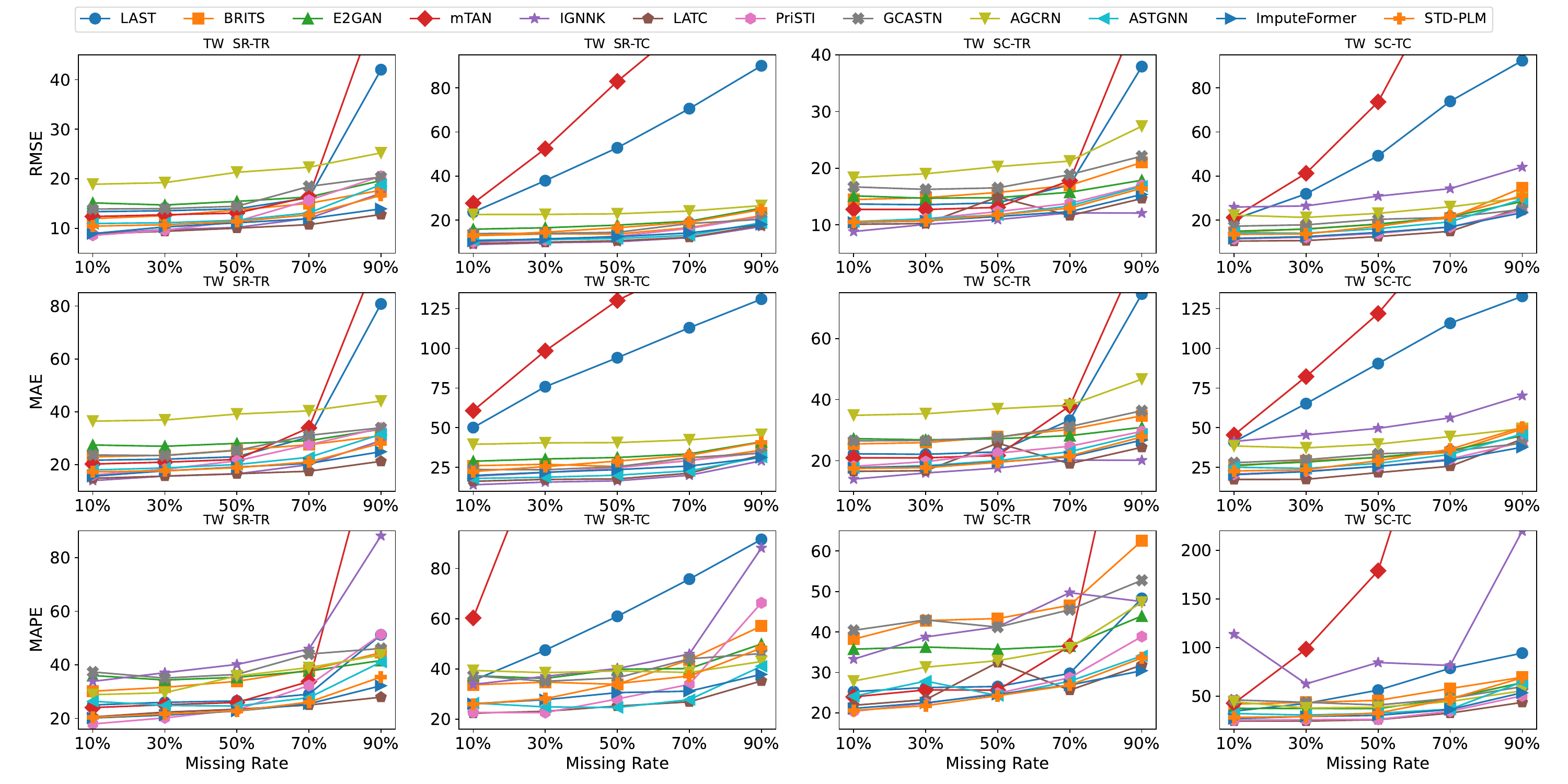}
    \caption{Performance of traffic data imputation for different models under 20 missing scenarios at TW.}
    \label{fig:TW315}
\end{figure*}

We report the imputation performance results of different models on the PEMS04, PEMS08, and Seattle datasets across four missing patterns and five missing rates. 
\begin{figure*}[htbp]
    \centering
    \begin{minipage}{\textwidth}
    \includegraphics[width=0.95\linewidth]{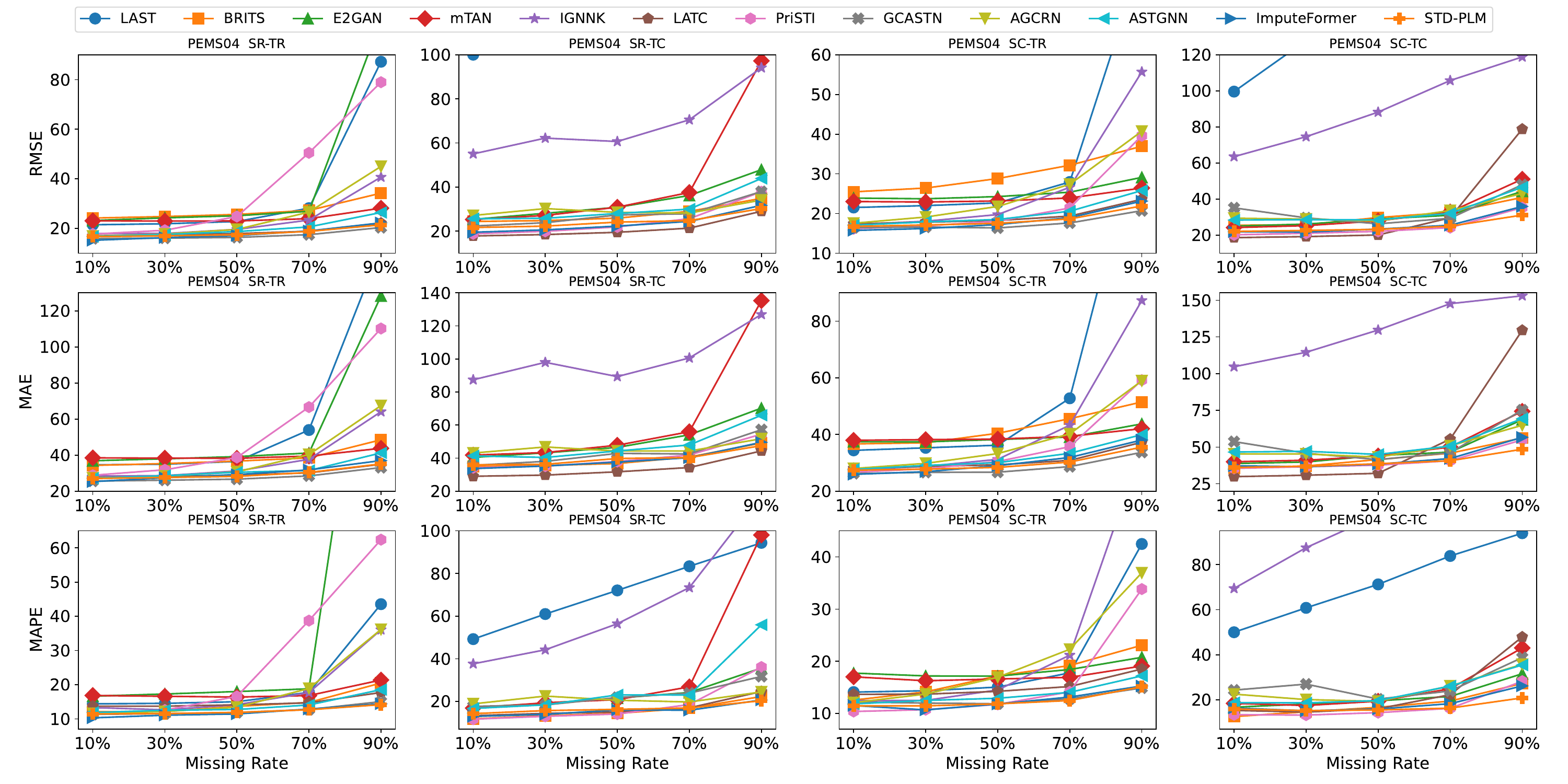}
    \caption{Performance of traffic data imputation for different models under 20 missing scenarios at PEMS04.}
    \label{fig:PEMS04}
    \end{minipage}
\vspace{0pt} 
    
    \begin{minipage}{\textwidth}
    \includegraphics[width=0.95\linewidth]{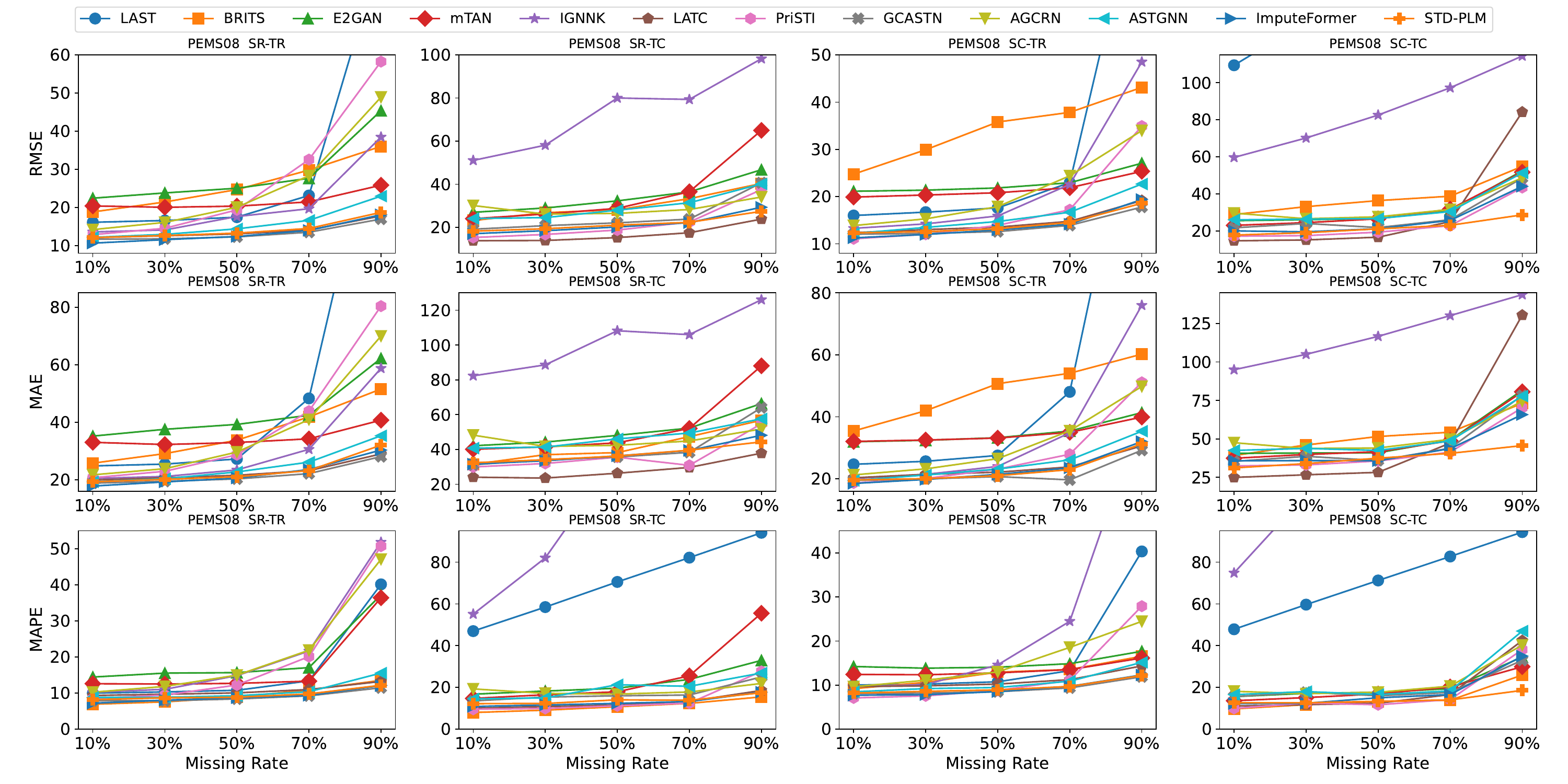}
    \caption{Performance of traffic data imputation for different models under 20 missing scenarios at PEMS08.}
    \label{fig:PEMS08}
    \end{minipage}
\end{figure*}
As shown in Figures~\ref{fig:PEMS04}, \ref{fig:PEMS08}, \ref{fig:Seattle}, and \ref{fig:TW315}, we observe that the performance declines as the missing rate increases. When the missing rate is below 0.5, the performance change of each model remains relatively stable. However, as the missing rate reaches 0.7 and 0.9, the performance of all models significantly deteriorates. 
This indicates that the noise of missing values at lower missing rates is minimal and does not substantially impact the model.
In contrast, when the missing rate exceeds 0.5, the abundance of missing values disrupts the inherent spatial-temporal relationships in the data. 



\textbf{ImputeFormer, GCASTN, BRITS, and STD-PLM} are the top-4 best deep-learning models for traffic data imputation, a common characteristic of their specialized design for handling missing data. ImputeFormer incorporates low-rankness into Transformers to capture spatiotemporal structures and mitigate the noise impact of missing value. GCASTN and BRITS use the time decay strategy to introduce time intervals between observed and missing values, while STD-PLM directly learns the spatial relationship of sensors by introducing the prior of spatial topology.
Therefore, efficiently incorporating prior knowledge of missing data may be a promising direction in traffic data imputation.

\textbf{BRITS, E$^2$GAN, mTAN.} These three models focus primarily on capturing temporal correlation. In the Seattle dataset, BRITS outperforms E$^2$GAN and mTAN, which can be attributed to its effective time decay strategy that captures small flow changes within local time. BRITS also reports well-performance on time continuous missing patterns in the PEMS04 and PEMS08 datasets, highlighting the suitability of time decay for modeling data trends. mTAN performs well with temporal random missing patterns by capturing global temporal relationships. It is neglected by BRITS and E$^2$GAN, which primarily emphasize local temporal correlations.

\textbf{IGNNK, PriSTI, GCASTN.} These three models account for spatial-temporal correlation. GCASTN excels in high missing rate scenarios (above 50\%) due to its contrastive learning approach, emphasizing data quality and model robustness. PriSTI delivers impressive results in low missing rate scenarios since it can deal with the missing value in advance and further consider their spatial-temporal dependence. IGNNK's underperformance stems from two factors: 1) the sparse adjacency matrice hinders spatial information messages, and 2) the lack of consideration for temporal information.


\begin{figure*}[h!]
    \centering
    \includegraphics[width=0.8
    \textwidth]{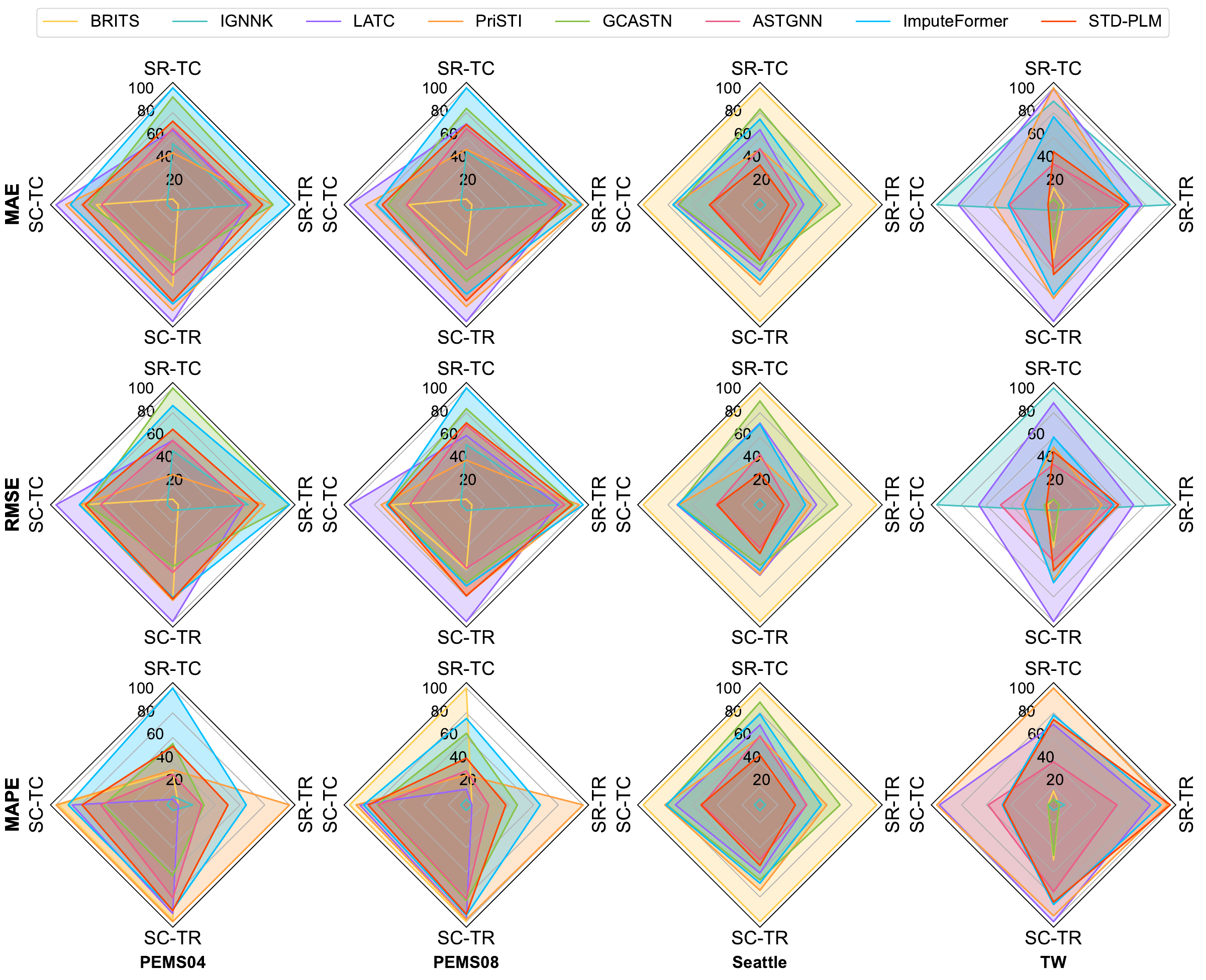}
    \caption{Comparison of models under different scenarios with low missing rates.} 
    \label{fig:low_missing_radar}
\end{figure*}

\textbf{LATC and ImputeFormer.} In contrast to deep learning models, LATC leverages the low-rank assumption of the matrix for model optimization. ImputeFormer combines low-rank with Transformer, further improving the performance. This indicates that taking into account the low-ranking features offers new insights on imputation tasks.


\begin{figure*}[h!]
    \centering
    \includegraphics[width=0.8
    \textwidth]{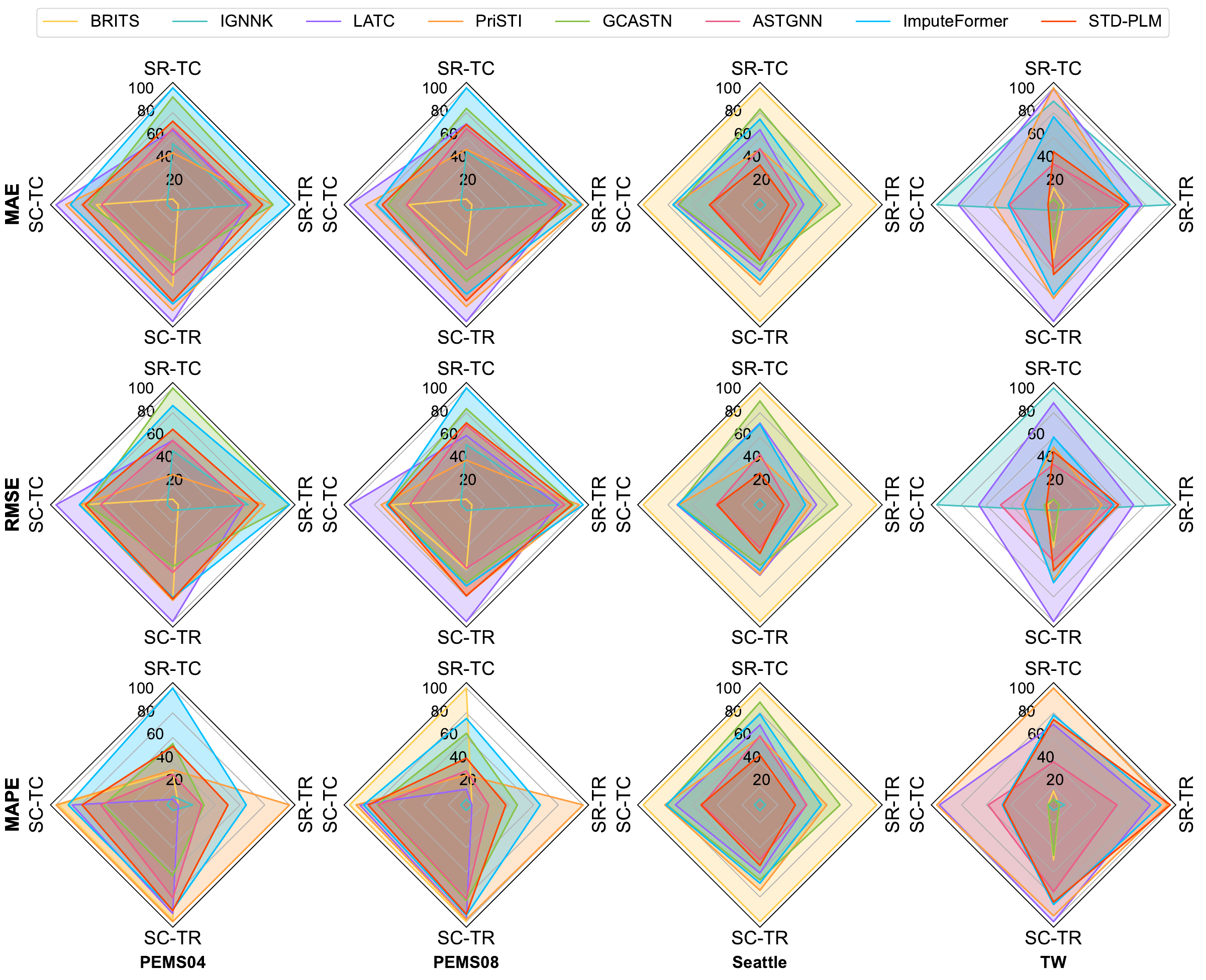}
    \caption{Comparison of models under different scenarios with high missing rates.} 
    \label{fig:high_missing_radar}
\end{figure*}

\begin{figure*}
	\centering
    \includegraphics[width=0.95\linewidth]{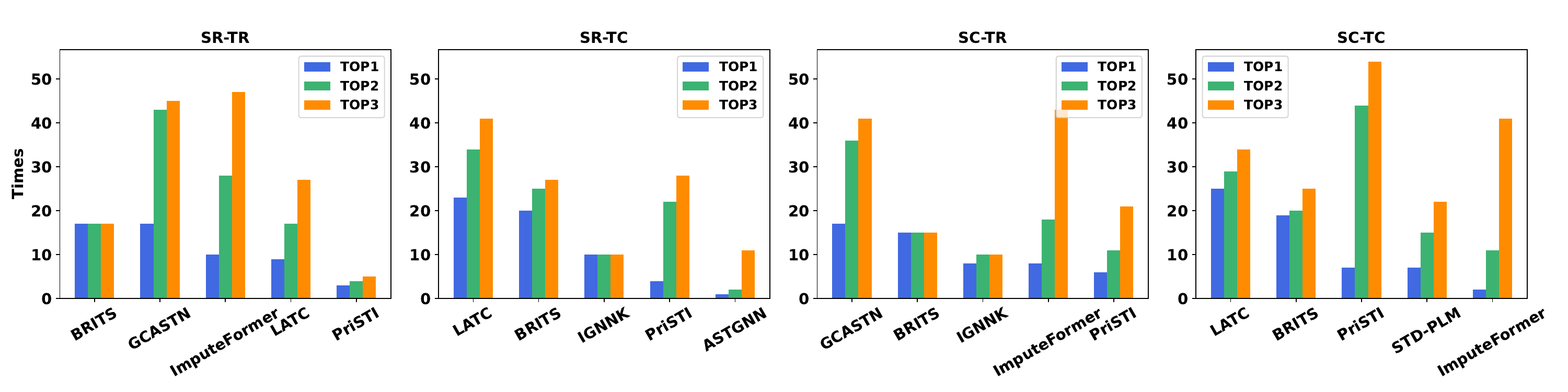}
	\caption{Model performance comparison at different missing patterns.}
	\label{fig:Missingpattern}
\end{figure*}

\subsubsection{Performance under Different Missing Scenarios}
{To comprehensively evaluate the overall performance of various models under different missing data scenarios, we categorize the missing rate into two levels: low missing rate ($< 50\%$) and high missing rate ($> 50\%$). Based on this categorization, we plot corresponding radar charts for each dataset and each evaluation metric to visualize model performance across different missing data settings, as shown in Figure~\ref{fig:low_missing_radar} and~\ref{fig:high_missing_radar}.}

{To make performance comparisons more intuitive, we select some of the well-performing models and normalize their metrics to a scale of [0, 100], where higher values indicate better results. Under this setting, the area enclosed by a model reflects its overall effectiveness, with larger areas representing stronger performance. Notably, the connections between two missing patterns on the radar charts also provide insights into a model’s capacity to handle hybrid missing patterns. For example, the line connecting SRTR and SCTR suggests that the missing mode is the mixing missing of both SRTR and SCTR.}

{We observe that dataset characteristics and missing data patterns significantly affect model performance. On the PEMS04 and PEMS08 datasets, ImputeFormer performs best under SR- missing patterns, while LATC shows superior results in SC- missing patterns. Both models share the use of tensor decomposition strategies, indicating that tensor decomposition is highly effective for handling missing data. Additionally, GCASTN and STD-PLM also demonstrate comparable performance.}

{For the Seattle traffic speed dataset, BRITS consistently outperformed other methods, and GCASTN is the second-best-performing model. This may be attributed to the relatively stable nature of speed over time, where temporal fluctuations are minimal. In such cases, a simple RNN with a time decay mechanism can effectively model the data, whereas more complex models might lead to performance degradation.}

{On the TW dataset, IGNNK achieves the best performance in most missing patterns, especially under TC- missing. This is mainly because the adjacency matrix of the TW dataset is relatively sparse, and the sensor values are similar to those of their neighboring sensors. Therefore, a simple GNN is sufficient to model the data effectively. However, under the SCTR missing pattern, IGNNK performed poorly due to the unavailability of signals from many nearby spatial sensors. In contrast, LATC considers global spatial relationships and achieves the best results in this case. Under the SCTC missing pattern, where most information is unavailable, models that rely on complex spatiotemporal relationships tend to perform worse.}

We also count how many times each model achieves the best, top-2, and top-3 performance. A higher count indicates stronger overall robustness and adaptability. As shown in Figure~\ref{fig:Missingpattern}, 
We observe that BRITS, GCASTN, PriSTI, ImputeFormer, and LATC achieve the top 3 best performances. A common characteristic of these models is the introduction of prior knowledge, where BRITS, GCASTN, and PriSTI utilize time delay, and ImputeFormer and LATC consider low-rank characteristics to capture global correlation. Yet other models only consider the spatio-temporal correlation, failing to efficiently impute, since extra noise of missing data leads to the spatiotemporal correlation being inaccurate. Thus, we guess that efficiently extracting extra prior knowledge from data may be a promising direction for more accurate imputation.

\begin{figure*}[t!]
	\centering
    \begin{minipage}[tb]{.32\linewidth}
            \subfigure{\scalebox{1}[1]{
		      \includegraphics[width=1\linewidth]{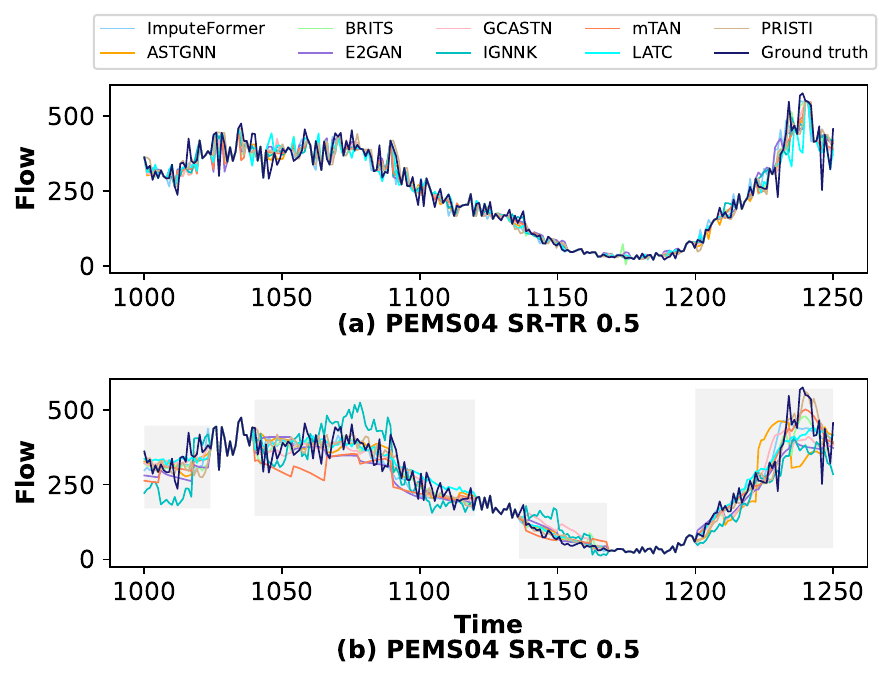}}
	    }
    \end{minipage}
\medskip
    \begin{minipage}[tb]{.66\linewidth}
        \subfigure{\scalebox{1}[1.2]{
		\includegraphics[width=0.48\linewidth]{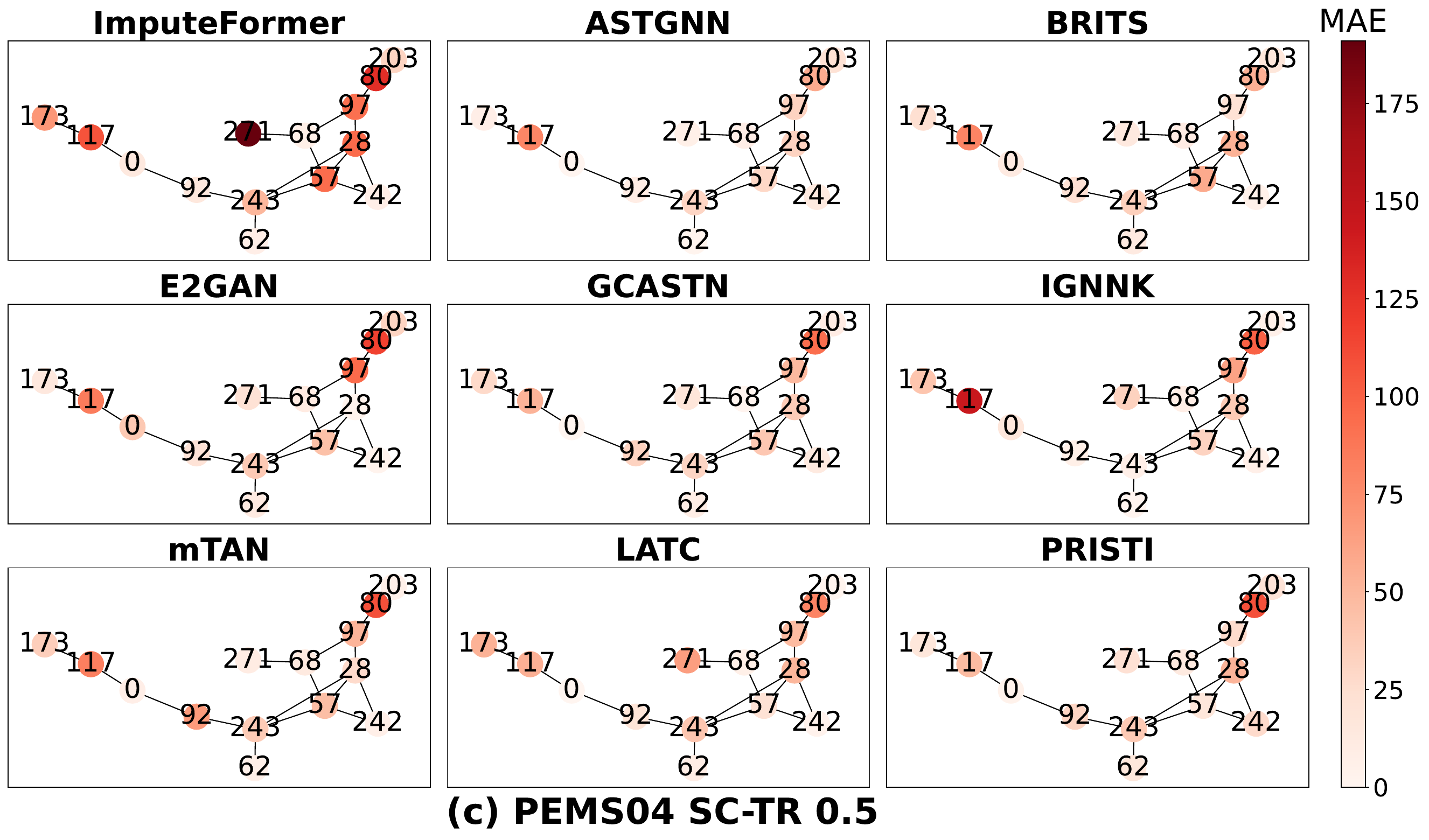}}
	}
	\subfigure{\scalebox{1}[1.2]{
		\includegraphics[width=0.48\linewidth]{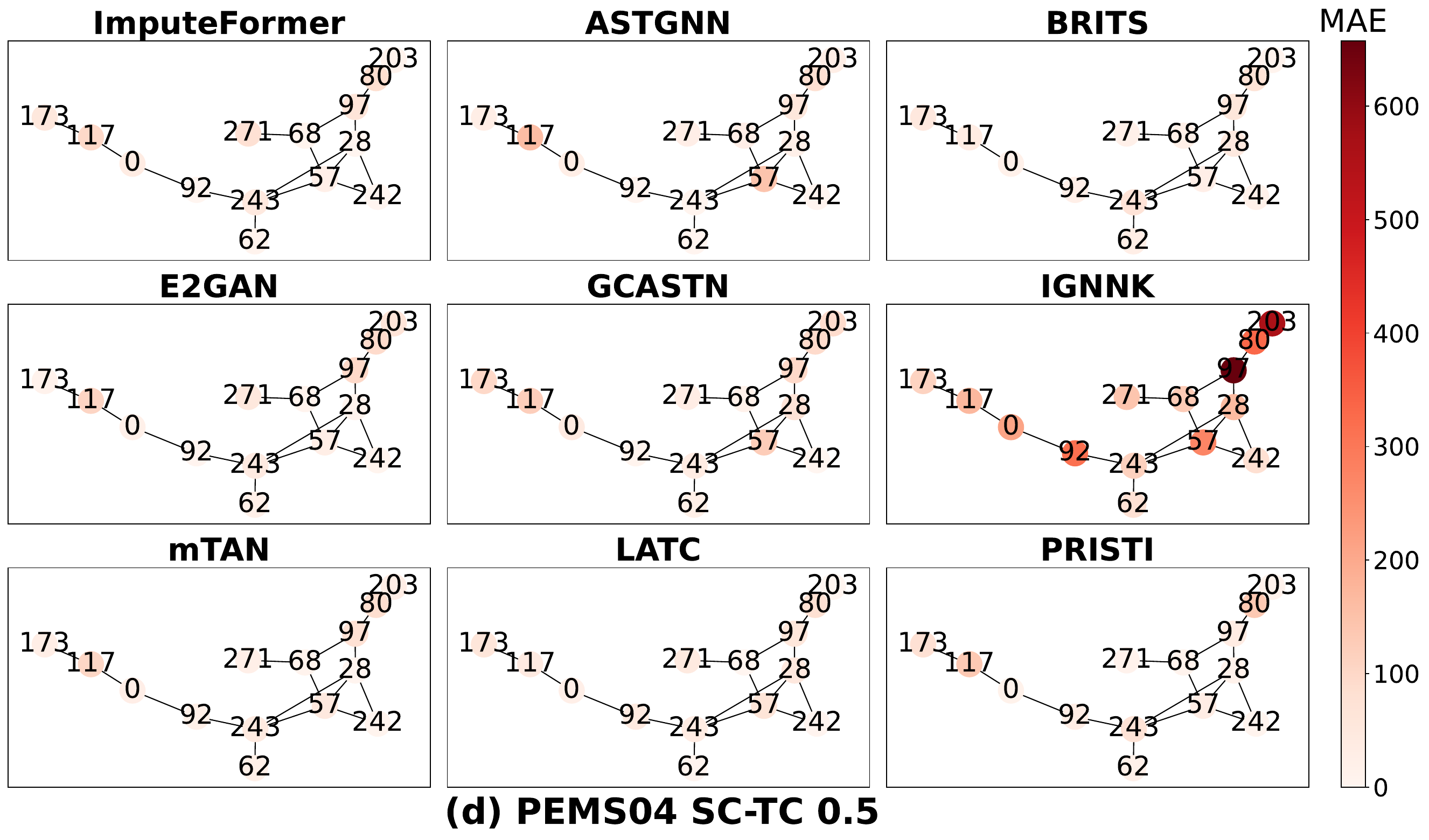}}
	}
    \end{minipage}
        \captionsetup{skip=-3pt}
	\caption{Visualization of model performance across different missing patterns with the missing rate of 0.5 on PEMS04.
}
	\label{fig:Casestudy}
\end{figure*}

\begin{table}[t]	
	\centering
	\caption{Performance comparison during challenging periods and stable periods.}
         \resizebox{1.0\columnwidth}{!}{
        \begin{tabular}{c|ccc|ccc}
            \toprule
            \multirow{2}{*}{Model}&\multicolumn{3}{c|}{Challenging Periods} &\multicolumn{3}{c}{Stable Periods}\\
            \cline{2-7}
            & MAE & RMSE & MAPE & MAE & RMSE & MAPE\\
            \hline
            LAST & 34.02 & 51.28 & 13.89 & 11.42 & 19.23 & 25.28\\
            BRITS & {25.64} & 40.95 & 10.25 & 10.21 & 17.22 & 26.84\\
            E2GAN & 33.37 & 50.46 & 13.18 & 11.73 & 18.42 & 30.38\\
            IGNNK & 26.82 & {39.48} & 10.71 & 8.54 & 13.34 & 22.30\\
            mTAN & 30.84 & 48.69 & 12.05 & 11.30 & 18.32 & 27.36\\
            PriSTI & 30.82 & 46.57 & 11.89 & 11.22 & 17.53 & 23.64\\
            GCASTN & \textbf{22.77} & \textbf{34.96} & \textbf{8.79} & \textbf{7.34} & \textbf{11.47} & \underline{18.30}\\
            AGCRN & 27.21 & 39.90 & 10.51 & 8.48 & 13.22 & 20.36\\
            ASTGNN & 26.53 & 42.62 & 9.96 & 8.21 & 13.22 & {20.34}\\
            LATC & 27.89 & 45.35 & 11.05 & 9.80 & 16.43 & 27.43\\
            ImputeFormer & {24.31} & {39.35} & {9.32} & \underline{7.46}& {12.52} & \textbf{17.34}\\
            STD-PLM & \underline{23.57} & \underline{35.38} & \underline{9.14} & 7.64 & \underline{12.00} & 18.63\\
            \bottomrule
        \end{tabular}
         }
	\label{tab:different_interval}
\end{table}

\subsubsection{Performance During Challenging and Stable Periods}
Most imputation models evaluate average performance in the test set, which fails to capture their strengths and weaknesses across different intervals. For instance, data imputation is more challenging during periods of fluctuating flow changes, while smoother flow transitions are easier to handle. To address this, we sort test samples by variance, with the top 25\% with the highest variance classified as challenging periods and the bottom 25\% as stable periods.
We report performance across these periods using the PEMS04 dataset with a 0.5 missing rate under the SRTR missing pattern, as shown in Table~\ref{tab:different_interval}. The results show that performance decreases significantly during challenging periods due to frequent fluctuations, whereas stable periods, characterized by smoother flow, exhibit more predictable temporal relationships.

By comparing models, GCASTN, STD-PLM, and ImputeFormer demonstrate superior performance in challenging and stable periods, respectively, highlighting the importance of spatial-temporal modeling. ImputeFormer effectively captures global knowledge and adapts to fluctuations. STD-PLM separately models the spatiotemporal relationships between nodes and leverages the language reasoning capabilities of LLM to learn traffic patterns in both challenging and stable periods, while GCASTN is well-suited for steady trends by utilizing the time decay strategy. BRITS and E$^2$GAN underperform as they focus solely on temporal dependencies, while spatial-temporal modeling methods such as ASTGNN and AGCRN produce more balanced performance.

\begin{table}[t]	
	\centering
	\caption{Time and memory cost. / notes that the model does not need an inference stage.}
         \resizebox{1.0\columnwidth}{!}{
        \begin{tabular}{c|ccc}
            \toprule
            \textbf{Model} & Training time (s) & Inference time (s) & Memory (MB) \\
            \hline
            BRITS & 2511.77 & 17.46 & 1612MB\\
            E$^2$GAN & 312.42 & 0.86 & 1513MB\\
            IGNNK & 632.00 & 1.59 & 4485MB\\
            mTAN & 955.49 & 4.22 & 1749MB\\
            PriSTI & 68762.25 & 8553.77 & 3195MB\\
            GCASTN & 138565.63 & 107.46 & 7594MB\\
            AGCRN & 7013.01 & 17.49 & 7579MB\\
            ASTGNN & 48539.37 & 84.74 & 6961MB\\
            LATC & 829.00 & / & 98MB\\
            ImputeFormer & 15320.95 & 8.19 & 5234MB\\
            STD-PLM & 12206.02 & 18.44 & 8744MB\\
            \bottomrule
        \end{tabular}
         }
	\label{tab:scabilitaty}
\end{table}

\subsection{Efficiency Evaluation}
We evaluate the training time, inference time, and memory usage of each model on the PEMS04 dataset with the SRTR missing pattern at a 0.5 missing ratio. Notably, the LATC model does not require inference. To ensure consistency, we adhere to the official code for all models, including the number of training epochs and early stopping settings. Detailed results are presented in Table~\ref{tab:scabilitaty}.

LATC has the smallest training time and memory usage. E2GAN, IGNNK, and mTAN also require less memory because their components are based on GRU or GNN with small computation costs. GCASTN, ImputeFormer, and ASTGNN utilize transformer architecture, so they demand more memory due to the quadratic space complexity of attention mechanisms. 
PriSTI exhibits significantly longer inference times. This is attributed to the diffusion model, which involves iterative denoising during inference and lacks parallelization support. Additionally, the need for multiple evaluations to estimate uncertainty further increases its inference time.

\subsection{Case study}
To more intuitively observe the difference between the imputation results of each model and the ground-truth, we conduct a case study for each missing pattern with a 0.5 missing rate at two different type datasets PEMS04 and Seattle. 

For the spatial random missing pattern, we randomly select a node to visualize the impute results and ground truth. As shown in Figure~\ref{fig:Casestudy} (a) and (b), we observe that the overall imputation curve of each model is aligned with the ground truth, demonstrating their ability to capture the temporal dependence.
Among them, BRITS, GCASTN, ImputeFormer, and LATC are more closely to real data. A common characteristic of these models is their incorporation of prior knowledge about missing data, which enhances imputation accuracy. BRITS and GCASTN leverage time-delay strategies to model temporal dependencies, while LATC and ImputeFormer adopt low-rank approaches to better capture the underlying data structure.

For the spatial continuous missing patterns, we select a cluster of missing nodes, visualize their connection relationships and observe the MAE metric of each node. The results as shown in Figure~\ref{fig:Casestudy} (c) and (d). At SCTR missing pattern, ASTGNN, GCASTN, BRITS, and PriSTI have smaller errors, indicating the importance of modeling temporal dependencies, especially when spatial information is unavailable. 
We also notice that  IGNNK performs poorly under the SCTC missing pattern due to its only focus on local spatial correlations and ignoring temporal dependencies, making it ineffective for this type of missing pattern.

\subsection{Summary of New Insights}
Through a large of experimental analysis, we find some interesting insights. 
\subsubsection{Suggestions for Model Design}
\leavevmode
\textbf{(1) Incorporating prior knowledge is usually beneficial.}
There are two common strategies for integrating prior knowledge into imputation models.
The first is to use prior knowledge to guide the imputation of missing values. For example, BRITS and GCASTN adopt a time-delay mechanism, while PriSTI first applies a linear imputation strategy. 
The second is to incorporate prior knowledge as a regularization term. For instance, LATC leverages a low-rank prior assumption and employs truncated nuclear normalization. 
Experimental results demonstrate that these models deliver strong and competitive performance.
\textbf{(2) Capture the global patterns such as periodicity, in traffic data is important.} We observe that tensor completion–based models, such as ImputeFormer and LATC, which exploit these global patterns, outperform most other models across most scenarios.

\subsubsection{Suggestions for Model Selection} 
\leavevmode
\textbf{(1) Considering different missing patterns,} GCASTN is the best model for temporal continuous missing pattern, and BRITS is the best model for spatial continuous missing pattern.
\textbf{(2) Considering memory-constrained scenarios,} TC is the best choice since LATC has the smallest memory cost. Among deep learning methods, PriSTI, ImputeFormer, and IGNNK strike a good balance between performance and memory efficiency. 
\textbf{(3) Considering the real-time requirements for inference speed.} TC has the fastest inference speed. For deep learning methods, E$^2$GAN is the fastest, and BRITS is the second fastest.

\subsubsection{Relations between Prediction and Imputation}
We find that spatial-temporal graph prediction models ASTGNN and AGCRN can achieve comparable performance in traffic imputation tasks, particularly in low missing rate scenarios. We think the reasons for this phenomenon is both of these two kinds of models focus on capturing the dynamics of traffic data along the spatial and temporal dimensions. Meanwhile, we can find that with the increase of missing rate, prediction models fail. This may because prediction model do not distinguish missing values and observed values, i.e., only treat them as different numbers. With the increase of missing rate, missing values may misguide models. Thus, as long as we can design a suitable mechanism to indicate missing values or use a proper missing initializing mechanism,  it is expected to unify these two tasks.

\section{Conclusion}\label{Sec:Conclusion}
In this paper, we propose a practice-oriented taxonomy of imputation models. Specifically, the models are systematically categorized based on their spatial–temporal modeling techniques and loss function designs. Furthermore, we develop a unified benchmarking pipeline to evaluate these models on four standardized traffic datasets under diverse missing patterns. Their performance is comprehensively assessed in terms of efficiency, effectiveness, and robustness across during challenging periods with significant traffic fluctuations.

Through extensive experiments, we derive several practical insights regarding model design, model selection under different scenarios, and the relation between imputation and prediction, which can provide valuable guidance for future research and real-world deployment.


\bibliographystyle{IEEEtran}
\bibliography{main}

\vfill
\end{document}